\documentclass{article}

\usepackage{PRIMEarxiv}

\usepackage[utf8]{inputenc} 
\usepackage[T1]{fontenc}    
\usepackage{hyperref}       
\usepackage{url}            
\usepackage{booktabs}       
\usepackage{amsfonts}       
\usepackage{nicefrac}       
\usepackage{microtype}      
\usepackage{lipsum}
\usepackage{fancyhdr}       
\usepackage{graphicx}       
\graphicspath{{media/}}     
\usepackage{amsmath}
\usepackage{tabularx}
\usepackage{algorithm} 
\usepackage{algpseudocode}
\usepackage{enumitem}
\usepackage{subcaption}

\title{Advancing Generative Model Evaluation: A Novel Algorithm for Realistic Image Synthesis and Comparison in OCR System
}

\author{
  Majid Memari\\
  School of Computing \\
  Southern Illinois University Carbondale \\
  IL 62901 USA  \\
  \texttt{memari@siu.edu} \\
    \And
    Khaled R. Ahmed \\
  School of Computing \\
  Southern Illinois University Carbondale \\
  IL 62901 USA  \\
  \texttt{khaled.ahmed@siu.edu} \\
   \And
  Shahram Rahimi \\
  Department of Computer Science and Engineering \\
  Mississippi State University \\
  Mississippi State, MS 39762 USA \\
  \texttt{rahimi@cse.msstate.edu} \\
     \And
  Noorbakhsh Amiri Golilarz \\
  Department of Computer Science and Engineering \\
  Mississippi State University \\
  Mississippi State, MS 39762 USA \\
  \texttt{amiri@cse.msstate.edu} \\
}

\begin{document}
\maketitle

\begin{abstract}
This research addresses a critical challenge in the field of generative models, particularly in the generation and evaluation of synthetic images. Given the inherent complexity of generative models and the absence of a standardized procedure for their comparison, our study introduces a pioneering algorithm to objectively assess the realism of synthetic images. This approach significantly enhances the evaluation methodology by refining the Fréchet Inception Distance (FID) score, allowing for a more precise and subjective assessment of image quality. Our algorithm is particularly tailored to address the challenges in generating and evaluating realistic images of Arabic handwritten digits, a task that has traditionally been near-impossible due to the subjective nature of realism in image generation. By providing a systematic and objective framework, our method not only enables the comparison of different generative models but also paves the way for improvements in their design and output. This breakthrough in evaluation and comparison is crucial for advancing the field of OCR, especially for scripts that present unique complexities, and sets a new standard in the generation and assessment of high-quality synthetic images.
\end{abstract}


\section{Introduction}
Handwritten text recognition, an integral part of computer vision and artificial intelligence, plays a vital role in a multitude of applications. Its importance is particularly evident in dealing with languages featuring complex scripts like Arabic, where the nuances in handwriting are notably diverse. The ability to digitize and recognize these intricate handwritten texts paves the way for advancements in fields such as automated translation, digital archiving, and efficient data retrieval from handwritten repositories \cite{castleman1996digital, niblack1985introduction, jain1989fundamentals, jahne2005digital, boyat2015review}.

Arabic Handwritten Digit Recognition, as a subset of OCR technology, stands as a significant challenge. The distinct and context-sensitive nature of Arabic numerals, varying in shape and form based on their positioning within a word, adds to the complexity of this task. This complexity is further compounded by variations in individual handwriting styles, writing tools, and the quality of paper used, making the accurate recognition of Arabic digits a formidable endeavor \cite{Al-Wzwazy2016HandwrittenNetworks, el2007two}.

The pursuit of enhancing Arabic Handwritten Digit Recognition's accuracy remains a key area of research, with deep learning, convolutional neural networks, and other machine learning algorithms at the forefront of these efforts. Despite advancements in OCR technology, challenges such as distorted, low-quality, or noisy text continue to impede the accurate recognition of handwritten text. Addressing these challenges through methods like data augmentation and real-time monitoring is essential for the improvement and evolution of OCR systems \cite{mori1999optical, mithe2013optical, islam2017survey, chaudhuri2017optical, singh2013optical, rao2016optical, berchmans2014optical}.

\subsection{Problem Statement}

A primary challenge in enhancing OCR technology, particularly for complex scripts like Arabic, is the subjective evaluation of generative models used for image synthesis. Although recent advancements have been made, OCR systems still grapple with the accurate interpretation of Arabic handwritten digits. The cursive and context-sensitive nature of Arabic numerals, which vary in shape and form based on their position in a word, adds a layer of complexity, making them more challenging to recognize than Latin numerals. Moreover, individual handwriting styles, writing tools, and paper quality further complicate the recognition process \cite{Al-Wzwazy2016HandwrittenNetworks}.

Deep learning, convolutional neural networks, and other machine learning algorithms are at the forefront of research aimed at enhancing Arabic Handwritten Digit Recognition. However, achieving high accuracy in OCR, especially for distorted, low-quality, or noisy text, remains a significant challenge. This is where the role of generative models becomes critical. Generative models are adept at producing a variety of synthetic images, but they often fall short in creating outputs that closely mimic the real intricacies of handwritten texts. This shortfall is particularly evident in scripts like Arabic, where the uniqueness of each character and its contextual form variations add to the complexity \cite{mori1999optical, mithe2013optical, islam2017survey, chaudhuri2017optical, singh2013optical, rao2016optical, berchmans2014optical}.

The core issue lies in the lack of an objective, standardized method to evaluate the realism and quality of images generated by these models. Traditional evaluation metrics often fail to capture the nuanced differences between synthetic and real handwritten texts, leading to a gap in the effective assessment and comparison of generative models. This gap hinders the development and optimization of models capable of producing high-fidelity synthetic images, crucial for training robust and accurate OCR systems. Our research aims to bridge this gap by proposing a novel algorithm that provides a more objective and nuanced evaluation of image realism in generative models.

\subsection{Research Objectives}
The central goal of this research is to enhance Optical Character Recognition (OCR) systems through novel methodologies, focusing particularly on Arabic handwritten digits. Our primary objective is the development of an innovative algorithm for the effective evaluation of generative models used in OCR. This algorithm seeks to establish a new standard for assessing the quality of synthetic images, emphasizing realism and relevance to real-world data.

Our specific objectives include:
\begin{itemize}
  \item Implementing data augmentation techniques to artificially expand and diversify the training dataset. This involves generating synthetic images with varying transformations and noise levels to enhance the robustness and accuracy of OCR systems \cite{namysl2019efficient, chernyshova2018generation, storchan2019data}.
  \item Introducing real-time monitoring to maintain an optimal balance between the quality and quantity of generated images. This approach allows the OCR model to learn effectively from augmented data without being compromised by poor-quality samples \cite{mori1992historical, chernyshova2018generation, kolak2003generative}.
  \item Addressing the instability in generative model training and implementing strategies to avoid issues such as mode collapse or vanishing gradients. This includes the application of real-time monitoring to provide insights into the model's performance and guide necessary adjustments \cite{kissos2016ocr, sporici2020improving}.
  \item Utilizing the insights from real-time monitoring to guide improvements in generative models, thereby enhancing their capacity to produce more realistic and diverse datasets for OCR training \cite{liu2023real, tkach2016sphere}.
  \item Establishing a benchmark for the comparison of different generative models and their hyperparameter settings, with a focus on their impact on OCR performance. This comparison will enable the selection of the most effective configurations for further refinement and optimization \cite{taghva1996evaluation, blue1994evaluation, cai2020real, fasel2005generative}.
\end{itemize}

Through these objectives, we aim to bridge the existing gap in generative model evaluation in the OCR domain, particularly for complex scripts like Arabic handwritten digits, setting a new direction for future advancements in the field.

\subsection{Scope and Significance}
The scope of our research extends into the domain of Optical Character Recognition (OCR) systems, with a special emphasis on enhancing the generation and evaluation of synthetic images. The cornerstone of our study is the development and implementation of data augmentation techniques and real-time monitoring, primarily focusing on Arabic handwritten digits. This is essential for constructing robust OCR systems capable of handling diverse input data, including low-quality or distorted images \cite{namysl2019efficient, chernyshova2018generation, storchan2019data}.

The significance of our research lies in its potential impact on advancing OCR technology, particularly for scripts that are historically challenging to digitize, like Arabic. We aim to establish:
\begin{itemize}
  \item \textbf{Advancement in OCR Technology}: Our approach significantly contributes to the development of advanced OCR systems by providing a more accurate and nuanced evaluation of synthetic image quality. This is crucial for scripts like Arabic, where the complexity and uniqueness of characters pose significant recognition challenges \cite{mori1999optical, mithe2013optical, islam2017survey, chaudhuri2017optical, singh2013optical, rao2016optical, berchmans2014optical}.
  \item \textbf{Enhancement of Image Generation}: We push the boundaries of synthetic image generation and evaluation, leading to the creation of more realistic and diverse datasets. This is vital for training robust OCR models capable of handling a wide range of real-world scenarios \cite{namysl2019efficient, chernyshova2018generation, storchan2019data}.
  \item \textbf{Benchmarking Generative Models}: Introducing a novel evaluation metric, we set a new benchmark in the field, allowing for a more objective comparison of generative models. This is key for driving innovation and excellence in model development \cite{cai2020real, fasel2005generative}.
  \item \textbf{Broader Implications}: The methodologies and principles developed in our study have potential applications beyond OCR, in various domains where image realism and quality are paramount. Thus, our research contributes to the broader field of computer vision and artificial intelligence.
\end{itemize}

Through these endeavors, our research offers substantial contributions to both the practical applications of OCR systems and the theoretical aspects of generative model evaluation. This includes addressing the instability in generative model training and implementing strategies to improve efficiency and accuracy, thereby setting a new direction for future advancements in the field \cite{mori1992historical, chernyshova2018generation, kolak2003generative, kissos2016ocr, sporici2020improving, liu2023real, tkach2016sphere, taghva1996evaluation, blue1994evaluation, cai2020real, fasel2005generative}.

\subsection{Structure of the Paper}
This paper is structured to provide a comprehensive understanding of the field and contribute to the advancement of Arabic Handwritten Digit Recognition models. The organization is as follows:

\begin{enumerate}
  \item \textbf{Introduction}: Provides a comprehensive background, outlines the problem statement, clarifies the research objectives, and discusses the scope and significance of the study.
  \item \textbf{Literature Review}: Delves into a detailed discussion on existing works related to Arabic Handwritten Digit Recognition, reviewing generative models in OCR and challenges in generating realistic images, as well as current methods of evaluating these models.
  \item \textbf{Methodology}: Explores methodologies and techniques employed in previous studies, their contributions to the field, and details the design and development of our novel evaluation algorithm, including its implementation within OCR systems.
  \item \textbf{Experimental Setup and Data Collection}: Describes the dataset used, criteria for model selection, specific parameters and conditions under which experiments were conducted, and compares the effectiveness of various techniques for image generation.
  \item \textbf{Results}: Presents findings from applying our algorithm, including a comparative analysis of different generative models and insights gained from the evaluation process.
  \item \textbf{Discussion}: Interprets the results, compares them with existing methods, discusses practical applications and limitations of our research, potential areas for future work, and critically examines the limitations and shortcomings of existing works in Arabic Handwritten Digit Recognition.
  \item \textbf{Conclusion}: Summarizes key findings and their implications, reinforcing the impact of our study on OCR technology and the field of image generation.
  \item \textbf{Acknowledgments}: Recognizes contributions and support from individuals, institutions, or funding bodies.
  \item \textbf{References}: Lists all academic sources and references cited throughout the paper.
\end{enumerate}

\section{Literature Review}

Generative models, particularly Generative Adversarial Networks (GANs) and Variational Autoencoders (VAEs), have significantly impacted Optical Character Recognition (OCR) technology. Their application in OCR systems is increasingly important for handling texts with high variability and complexity, such as Arabic scripts.

\textbf{Generative Adversarial Networks (GANs)}: These models, known for their dual-network architecture, have been instrumental in generating synthetic images for OCR training. By creating diverse datasets, GANs help in developing robust models with enhanced accuracy and generalization capabilities \cite{Goodfellow2014, bond2021deep}.

\textbf{Variational Autoencoders (VAEs)}: VAEs excel in generating new data instances by learning the distribution of input data. In OCR, VAEs are crucial for creating synthetic data that mimic complex script characteristics, including cursive and context-sensitive scripts like Arabic \cite{kingma2019introduction, girin2020dynamical}.

\textbf{Conditional Variants}: C-VAE and C-GAN, the conditional extensions of VAEs and GANs, respectively, stand out for their ability to generate high-quality synthetic images. These models can produce images with specific desired attributes, enhancing OCR applications \cite{Sohn2015CGAN, Mirza2014, lim2018molecular, asperti2020balancing}.

Despite their effectiveness, evaluating these generative models remains a challenge. Traditional metrics often fail to accurately assess the quality and realism of generated images, particularly in the context of complex scripts. This gap in evaluation highlights the need for advanced, objective methods to assess these models' performance in OCR scenarios \cite{aggarwal2021generative, cetin2023attri, doermann2014handbook, cai2019multi, kebaili2023deep}.

In summary, while generative models like GANs and VAEs have revolutionized OCR, their full potential hinges on the development of more refined evaluation methods. This subsection underlines the importance of innovative approaches in assessing and comparing generative models, particularly in advancing OCR technology through improved evaluation metrics.

\subsection{Challenges in Realistic Image Generation}

Generating realistic synthetic images is a paramount challenge faced by generative models in Optical Character Recognition (OCR). The aim is not only to produce visually convincing images but to retain essential characteristics of the target data, like handwriting in various scripts. This is particularly daunting for languages with complex scripts, such as Arabic, where nuances like stroke variation, character connectivity, and contextual shaping require high fidelity in synthetic image generation.

The core difficulty lies in the model's capacity to accurately replicate natural handwriting's subtleties and variations. Each script and handwriting style introduces unique patterns and idiosyncrasies, demanding precise learning and mimicry by generative models. Factors such as writing tool variations, paper texture, and environmental impacts on the writing process further complicate this task.

Moreover, defining 'realism' in synthetic images is subjective, and quantitative metrics may not fully align with human perception of realism. This discrepancy presents a significant challenge in evaluating generative models, as the goal is to produce data indistinguishable from real handwriting, especially in OCR applications where accuracy and authenticity are crucial.

Additionally, there's a risk of overfitting, where models might generate images too closely aligned with training data, leading to a lack of diversity and generalizability in synthetic images. Balancing realism with diversity to enhance OCR systems becomes a critical aspect of research in this area.

This subsection outlines the multifaceted challenges in generating realistic synthetic images, underscoring the need for sophisticated generative models and evaluation methods to effectively address these complexities.

\subsection{Current Evaluation Methods}

Evaluating generative models in OCR, especially for complex scripts, involves various metrics, each with specific applications and inherent limitations. Understanding these metrics is crucial to identify the gaps that our proposed fine-tuning of the FID score aims to fill.

\textbf{Mean Squared Error (MSE)} and \textbf{Peak Signal-to-Noise Ratio (PSNR)} are traditional metrics focused on pixel-level accuracy. MSE measures the average squared difference between pixels of generated and real images, while PSNR compares the similarity based on pixel value noise levels. However, both fail to capture perceptual aspects of image quality, crucial for the realism of handwritten texts.

\textbf{Structural Similarity Index (SSIM)} advances evaluation by considering changes in texture, luminance, and contrast, offering a more comprehensive assessment than MSE and PSNR. Despite this, SSIM does not entirely capture the handwriting's stylistic nuances, critical for OCR accuracy.

\textbf{Fréchet Inception Distance (FID)} score represents a significant advancement in evaluating generative models. FID measures the distance between feature vectors of real and generated images within an embedded space, thus assessing both the diversity and quality of generated images. While effective in many contexts, FID's standard application may not fully align with specific characteristics of datasets like Arabic handwritten digits, potentially leading to skewed assessments.

\textbf{Inception Score (IS)} and \textbf{Mode Score} also provide insights into the quality and diversity of generated images. IS evaluates image clarity and diversity based on the predictive power of a pre-trained Inception model, whereas Mode Score extends this by considering the diversity of generated data relative to real data. However, both scores have limitations in capturing the fidelity of generated images to real data, especially in handwriting recognition.

The \textbf{Kernel Inception Distance (KID)} is another metric that measures similarity between sets of images by comparing Inception embeddings. KID, unlike FID, is unbiased and provides more reliable results, especially for smaller sample sizes. However, it might not fully encapsulate the complexity of handwriting in OCR.

Each of these metrics, while valuable, exhibits gaps in effectively evaluating the realism and applicability of synthetic images in OCR. Our research focuses on fine-tuning the FID score to make it more relevant and meaningful for specific datasets like Arabic handwritten digits. This fine-tuning aims to address the unique challenges presented by complex scripts and improve the overall effectiveness of generative models in OCR.

Table \ref{tab:generative_models} summarizes the state of the art of image generation techniques in improving OCR performance. These methods offer various advantages, such as handling noise and distortion or generating high-quality, diverse images. However, they also have limitations, such as training instability, blurry image generation, or resource requirements.

\begin{table*}[ht] 
  \caption{Generative Models}
  \label{tab:generative_models}
  \begin{tabularx}{\textwidth}{@{}lXXX@{}}
    \toprule
    Model & Description & Advantages & Limitations \\
    \midrule
    Variational Autoencoders (VAEs) \cite{Kingma2013} & A generative model that learns a latent representation of the input data, generating new samples. & Effective at modelling complex data distributions. Capable of handling noise and distortion. & Can produce blurry images. Less control over generated image features. \\
    \midrule
    Generative Adversarial Networks (GANs) \cite{Goodfellow2014} & A framework where a generator and discriminator compete to improve the generated image quality. & Generate high-quality, sharp images. Effective at generating diverse images. & Training can be unstable. Can suffer from mode collapse (limited variety in samples). \\
    \midrule
    Cycle-Consistent Adversarial Networks (CycleGANs) \cite{Zhu2017Cycle} & An unsupervised image-to-image translation technique that leverages cycle consistency. & Enables unsupervised learning for image translation. Can generate visually consistent images. & May struggle with complex transformations. Limited to paired domain mappings. \\
    \midrule
    StyleGAN (and its variants) \cite{karras2019style} & A GAN architecture that allows control over the generated image's style, content, and stochastic features. & Enables fine-grained control over image generation. Generates high-quality images. & Requires large datasets and computational resources. Potential overemphasis on style. \\
    \midrule
    BigGAN \cite{brock2018large} & A GAN architecture that scales up in capacity and size, resulting in high-quality images. & Generates high-resolution, high-quality images. Capable of generating diverse images. & Requires large datasets and computational resources. Can suffer from mode collapse. \\
    \bottomrule
  \end{tabularx}
\end{table*}

\subsection{Evaluation Metrics for Generative Models}
Evaluation metrics play a crucial role in assessing the performance of generative models, allowing us to measure the quality, diversity, and fidelity of the generated samples. These metrics are designed to provide quantitative insights into how well the models capture the characteristics of the real data distribution. 

Evaluation metrics for generative models can be broadly categorized into two groups shown in the figure \ref{fig:evaluation}:

\begin{figure*}[htbp]
    \centering
    \includegraphics[width=\textwidth]{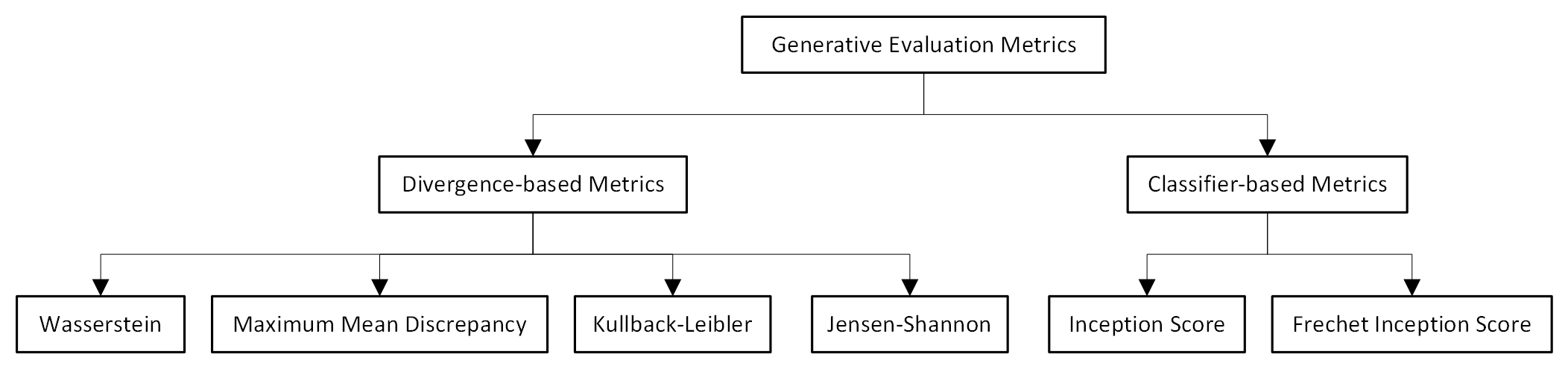}
    \caption{Evaluation Metrics for Generative Models}
    \label{fig:evaluation}
\end{figure*}

\subsubsection{Classifier-based Metrics}
These metrics utilize pre-trained classifiers to extract features from both real and generated samples, enabling a comparative analysis. Two commonly used metrics in this category are the Inception Score (IS) and the Fréchet Inception Distance (FID).

\textbf{Inception Score (IS)}
Measures the quality and diversity of generated samples by evaluating both the conditional and marginal distributions. It utilizes a pre-trained Inception model to extract features and provides an aggregate score that combines these two aspects \cite{barratt2018note}.

The IS measures these two aspects as follows:
\begin{itemize}
    \item \textbf{Diversity}: Good models should produce a variety of different images, not just variations of the same image. This is quantified by the entropy of the marginal distribution over labels. A higher entropy means more diversity.
    \item \textbf{Quality}: Good models should generate images that look like the training data. This is quantified by the conditional distribution of the labels given the generated images. If a model is good, the conditional distribution should have low entropy for each image (i.e., the model is confident about the label of the image).
\end{itemize}
Let's denote:
\begin{itemize}
    \item $p(y|x)$ as the conditional distribution of the label $y$ given the image $x$, produced by the Inception model.
    \item $p(y)$ as the marginal distribution over labels, obtained by integrating over all images $x$ in the generated data: $p(y) = \int p(y|x) dx$.
\end{itemize}

The Inception Score is then defined as:
\begin{equation}
IS = \exp\left( \mathbb{E}_x \left[ \text{KL}(p(y|x) || p(y)) \right] \right) \text{,}
\end{equation}
where $\mathbb{E}_x$ denotes the expectation over all generated images $x$, and $\text{KL}$ is the Kullback-Leibler divergence, which measures how one probability distribution is different from a second, reference probability distribution.

\textbf{Fréchet Inception Distance (FID)}
quantifies the distance between the feature representations of real and generated samples, leveraging a pre-trained Inception model. The FID score provides a measure of similarity between the distributions and is particularly useful for assessing the overall fidelity and diversity of generated samples.

The Fréchet distance is a metric for measuring the similarity between two curves. Mathematically, for two continuous curves $C1: [0,1] \rightarrow M$ and $C2: [0,1] \rightarrow M$ in a metric space $(M,d)$, it's defined as:
\begin{equation}
F(C1,C2) = \inf_{\alpha, \beta} \max_t d(C1(\alpha(t)),C2(\beta(t))), \text{ for all } t \in [0,1]
\end{equation}

The Fréchet Inception Distance (FID) score extends this concept to compare the distributions of real and generated images in the feature space of a pre-trained Inception model. The steps to calculate the FID score include preprocessing the images, passing them through the Inception model, and calculating the mean and covariance of the activations for both real and generated images. The FID score is then computed as:
\begin{equation}
FID = d^2 + \text{trace} ,
\end{equation}
where $d^2 = \left|\left|\mu_{\text{real}} - \mu_{\text{gen}}\right|\right|^2$ is the squared Euclidean distance between the means. $\text{trace} = \text{Tr}(\Sigma_{\text{real}} + \Sigma_{\text{gen}} - 2(\Sigma_{\text{real}}\Sigma_{\text{gen}})^{1/2})$ is the trace of the sum of covariance matrices and their square root. A lower FID score indicates the generated images are of higher quality and more similar to the real ones.

Although we have provided an overview of its calculation here, a more comprehensive discussion of the FID score, particularly its fine-tuning to better suit specific datasets, will be elaborated in the Methodology chapter. This forthcoming section will provide detailed insights into how we adapt and apply the FID score in the context of our specific research and data. By optimizing this measure for our unique dataset, we aim to achieve a more precise evaluation of the performance of our generative models. Stay tuned for this in-depth exploration in the upcoming chapters.

\subsubsection{Divergence-based Metrics}
These metrics estimate divergences or distances between the distributions of real and generated samples, providing a more principled measurement of fidelity and diversity. Some commonly used metrics in this category include Kullback-Leibler divergence (KL), Jensen-Shannon divergence (JS), Wasserstein distance (W), and Maximum Mean Discrepancy (MMD).

\textbf{Kullback-Leibler Divergence (KL):}
Measures the divergence between the distributions of real and generated samples. It quantifies the difference in information content between the two distributions \cite{contreras2012kullback}.
The KL Divergence between two probability distributions $P$ and $Q$ is defined as:
\begin{equation}
KL(P || Q) = \sum P(x) \log \left(\frac{P(x)}{Q(x)}\right) \quad \text{for all } x
\end{equation}
It measures the difference in the information content between the two distributions, $P$ and $Q$.

\textbf{Jensen-Shannon Divergence (JS):}
Calculates the divergence between two probability distributions, typically the real and generated data distributions. It captures the similarity and difference between the two distributions \cite{menendez1997jensen}.

The JS Divergence is another method to measure the similarity between two probability distributions, $P$ and $Q$. It's defined as:
\begin{equation}
JS(P || Q) = \frac{1}{2} KL(P || M) + \frac{1}{2} KL(Q || M)
\end{equation}
where \( M \) is the average of \( P \) and \( Q \), defined as \( M = \frac{1}{2} (P + Q) \).

\textbf{Wasserstein Distance (W):}
Also known as Earth Mover's Distance, it measures the distance between the real and generated data distributions by computing the minimum cost of transforming one distribution into another \cite{vallender1974calculation}.

The Wasserstein distance between two probability distributions $P$ and $Q$ is defined as the solution of the following optimization problem:
\begin{equation}
W(P,Q) = \inf \sum |x_i - y_i| \cdot P(T=i) \text{ for all } i
\end{equation}
where the infimum is taken over all joint distributions of $(X, Y)$ with marginal distributions $P$ and $Q$.

\textbf{Maximum Mean Discrepancy (MMD):}
Measures the distance between the means of the real and generated data distributions in a reproducing kernel Hilbert space. It provides a principled way to measure the similarity between the two distributions \cite{borgwardt2006integrating}.

The MMD between two distributions $P$ and $Q$ in a reproducing kernel Hilbert space $H$ with a kernel $k$ is defined as:
\begin{equation}
MMD(P,Q) = \sup \left| \left| E_P [k(X,.)] - E_Q [k(Y,.)] \right| \right|_H
\end{equation}
where $\sup$ denotes the supremum, $E_P$ and $E_Q$ denote the expectations under the distributions $P$ and $Q$, $X$ and $Y$ are random variables with distributions $P$ and $Q$ respectively, and $||.||_H$ denotes the norm in the Hilbert space $H$.

Divergence-based metrics offer a more rigorous and theoretically grounded approach but may be computationally expensive and require access to the true data distribution or its samples. When selecting evaluation metrics for generative models, researchers must consider factors such as the specific characteristics of the data, research goals, computational efficiency, and interpretability. Understanding the strengths and limitations of different metrics is essential to choose the most appropriate evaluation approach for a given scenario.
Table \ref{tab:evaluation_metrics} provides a comparative analysis for various evaluation metrics for generative models:

\begin{table*}[ht]
\caption{Evaluation Metrics for Generative Models}
\label{tab:evaluation_metrics}
\begin{tabularx}{\textwidth}{@{}lXXX@{}}
\toprule
\textbf{Metric} & \textbf{Methodology} & \textbf{Pros} & \textbf{Cons} \\ \midrule
Inception Score (IS) \cite{barratt2018note}
& Uses a pre-trained Inception model to extract features from real and generated samples, compares the conditional and marginal distributions of generated data.
& Measures both the realism and diversity of the generated samples.
& Sensitive to the choice of classifier, potential bias towards certain types of data, does not account for issues like class-conditional generation or memorization. \\ \midrule
Frechet Inception Distance (FID) \cite{Heusel2017}
& Measures the distance between real and generated data distributions in the feature space of an Inception network.
& Provides a holistic view of the quality and diversity of generated samples, better correlation with human judgement than IS.
& Assumes high-level and complex features in images, may not be effective for low-quality images. \\ \midrule
Kullback-Leibler Divergence (KL) \cite{contreras2012kullback}
& Estimates the divergence between real and generated data distributions.
& Provides a principled measurement of the fidelity and diversity of the generated samples.
& Computationally expensive, requires access to the true data distribution or its samples, difficult to interpret or compare across different models or datasets. \\ \midrule
Jensen-Shannon Divergence (JS) \cite{menendez1997jensen}
& Calculates the divergence between two probability distributions, typically between the real and generated data distributions.
& Symmetric measure, provides a principled measurement of the similarity between real and generated samples.
& Computationally expensive, requires access to the true data distribution or its samples, difficulty to interpret or compare across different models or datasets. \\ \midrule
Wasserstein Distance (W) \cite{vallender1974calculation}
& Also known as the Earth Mover’s Distance, measures the distance between the real and generated data distributions.
& Provides robust and principled ways to measure the fidelity and diversity of the generated samples.
& Computationally expensive, requires a lot of resources and time, could be difficult to interpret. \\ \midrule
Maximum Mean Discrepancy (MMD) \cite{borgwardt2006integrating}
& Measures the distance between means of the real and generated data distributions in a reproducing kernel Hilbert space.
& Does not require density estimation, provides a principled way of measuring the similarity between two distributions.
& Difficult to choose an appropriate kernel, requires access to the true data distribution or its samples, can be computationally intensive. \\
\bottomrule
\end{tabularx}
\end{table*}

As our research focuses specifically on real-time optical character recognition (OCR), it becomes increasingly important to find an evaluation metric that not only accurately measures the performance of generative models but also does so quickly. Optical character recognition is a time-sensitive application where rapid generation and evaluation of synthetic images are crucial for efficient operation. 

Handwritten Optical Character Recognition (OCR) is a challenging task due to the high degree of variation in handwriting. Handwriting varies significantly from person to person, and even the same person's handwriting can change over time or under different conditions. It can also be affected by factors such as the writing instrument used, the writing surface, and the speed at which the person is writing. This makes it more difficult for OCR systems to correctly identify characters, as there is no single 'template' that can be used to match each character.

Image generation, particularly through Generative Adversarial Networks (GANs) and Variational Autoencores (VAEs), has the potential to aid in this task. Generative models can be trained to generate a wide range of examples for each character, which can help an OCR system to learn the diversity of appearances each character can have. This can lead to a more robust OCR system that is better able to handle the variability in handwriting.

However, the evaluation metrics for generative models, which are often used to assess the quality of generated images, have limitations when it comes to recognizing low-quality images. These metrics, such as the Inception Score (IS), Frechet Inception Distance (FID), Kullback-Leibler Divergence (KL), and others, are designed to measure the overall quality and diversity of the generated images, rather than focusing on specific details that might be crucial for OCR \cite{Heusel2017,ahmed2016improving}.

For example, a model might generate images that score highly on these metrics, indicating that they have high quality and diversity, but if the images are not high-resolution or clear enough, the OCR system may still struggle to correctly identify characters.

Furthermore, these metrics can be computationally intensive, which makes it harder to use them in real-time applications or when dealing with large volumes of data.
Therefore, while image generation can potentially improve the performance of handwritten OCR systems, there is still a need for improved evaluation metrics that can accurately assess the quality of generated images in the context of OCR, particularly when dealing with low-quality images or real-time applications.

In this chapter, we conducted a thorough examination of the literature concerning Generative Models designed to enhance Optical Character Recognition (OCR) performance, as well as the assessment metrics employed for these models. In the following chapter, we will delve into the specific Generative Models we deployed to augment the Arabic Handwritten Digit Dataset in OCR. Additionally, we will introduce our innovative method for accurately calculating the Fréchet Inception Distance (FID) score, providing a robust measure for the quality of the images generated.

\section{Methodology}
The purpose of this methodology chapter is to provide a detailed and comprehensive explanation of the research design, data collection, and analysis methods employed in this study. These methodological components are essential for understanding the process and rationale behind the investigation into the effectiveness of Conditional Generative Adversarial Networks (C-GANs) and Conditional Variational Autoencoders (C-VAEs) for enhancing Optical Character Recognition (OCR) performance in Arabic handwritten digit recognition.

In this study, we aim to compare C-GANs and C-VAEs for image generation, introduce a new approach to correctly calculate the FID score for monitoring the quality of generated images. We also propose another new approach to examine the performance of the improved OCR systems using Saliency Maps. The research design, data collection, and analysis methods will guide the investigation and ensure that the results can provide valuable insights into the use of generative models for OCR performance enhancement.

\subsection{Research Design}
The overall research design of this study is a comparative, quantitative, and experimental approach. This design was chosen because it allows for a systematic comparison of C-GANs and C-VAEs in generating synthetic images and enhancing OCR performance. By employing an experimental approach, we can measure and analyze the impact of different generative models on OCR systems and identify the most effective strategies for improving accuracy and efficiency.
The choice of using GANs and VAEs for data augmentation in OCR systems is based on their proven effectiveness in quickly generating synthetic images and their potential for improving the performance of machine learning models. Both C-GANs and C-VAEs have been successfully applied to a wide range of computer vision tasks, including image synthesis, style transfer, and image inpainting, making them relevant and promising candidates for enhancing OCR performance \cite{apuke2017quantitative}.

\subsection{Dataset}
The dataset used for this study is the Arabic Handwritten Digits Dataset (AHDD), which was was created by El-Sawy et al. \cite{Sawy2017} The dataset contains 60,000 grayscale images of handwritten Arabic digits, with each digit having 6,000 samples. The images are 28x28 pixels in size and represent Arabic digits ranging from 0 to 9 which is shown below in figure \ref{fig:real}. This dataset is characterized by its variability in writing styles, as well as the presence of noise and distortions, which makes it a suitable choice for investigating the effectiveness of GANs and VAEs in improving OCR systems.
\begin{figure}
    \centering
    \includegraphics[width=1\linewidth]{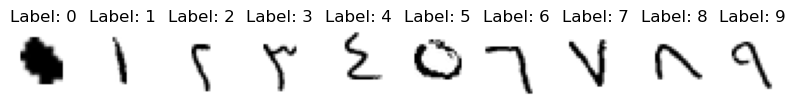}
    \caption{Arabic Handwritten Digits Dataset (AHDD)}
    \label{fig:real}
\end{figure}
The dataset was divided into three subsets: training, validation, and test sets. Following the standard practice in machine learning research, 70\% of the data was allocated to the training set, 15\% to the validation set, and 15\% to the test set [35]. This division ensures that the models are trained on a sufficiently large dataset while providing separate subsets for model selection and evaluation.

\subsection{Synthetic Image Generation}
The process of generating synthetic images using C-GAN and C-VAE involved several steps, including pre-processing and hyperparameter tuning. The pre-processing steps included data normalization, which involved scaling the pixel values to a range between 0 and 1, and data augmentation through random transformations, such as rotation and scaling. For both C-GAN and C-VAE, various architectures and hyperparameter settings were explored to optimize the quality of the generated images. Hyperparameters included learning rates, batch sizes, and the number of training epochs \cite{cubuk2018autoaugment}.

\subsubsection{C-GAN}
The algorithm of the Conditional Generative Adversarial Network (C-GAN) \cite{Mirza2014} , as shown in figure 3, is an extension of the GANs architecture, which was initially proposed by Goodfellow et al. \cite{Goodfellow2014}. GANs consist of two neural networks: a generator that creates synthetic data samples and a discriminator that distinguishes between real and synthetic samples. These two networks are trained simultaneously in a min-max game, where the generator tries to create samples that can fool the discriminator, while the discriminator attempts to accurately classify the samples as real or synthetic.

C-GANs incorporate additional conditional information (e.g., class labels) into both the generator and discriminator networks, allowing the model to generate samples based on specific conditions. The conditional information is typically concatenated with the input noise vector for the generator and with the data samples for the discriminator. This modification enables the generation of more targeted and diverse samples, making C-GANs particularly useful for various applications, such as data augmentation and image synthesis \cite{frid2018synthetic}.

\begin{figure*}[ht]
    \centering
    \begin{minipage}{.45\textwidth}
        \begin{algorithm}[H]
            \caption{Conditional GAN}
            \linespread{0.9}\selectfont 
            \label{alg:conditional_gan}
            \begin{algorithmic}[1]
                \State Initialize $G$ and $D$ with random weights
                \State Set hyperparameters: $\alpha$, $T$, $B$
                \For{$t = 1$ to $T$}
                    \For{mini-batch $x$, $y$}
                        \State Sample noise $z$
                        \State Generate fake samples $\hat{x} = G(z, y)$
                        \State Update the discr.: compute $\mathcal{L}_D$ and update $\theta_D$
                        \State Sample new noise $z$
                        \State Generate new fakes $\hat{x} = G(z, y)$
                        \State Update the gen.: compute $\mathcal{L}_G$ and update $\theta_G$
                    \EndFor
                \EndFor
            \end{algorithmic}
        \end{algorithm}
        \caption{The Conditional GAN Algorithm}
    \end{minipage}
    \hfill
    \begin{minipage}{.45\textwidth}
        \centering
        \includegraphics[width=\linewidth]{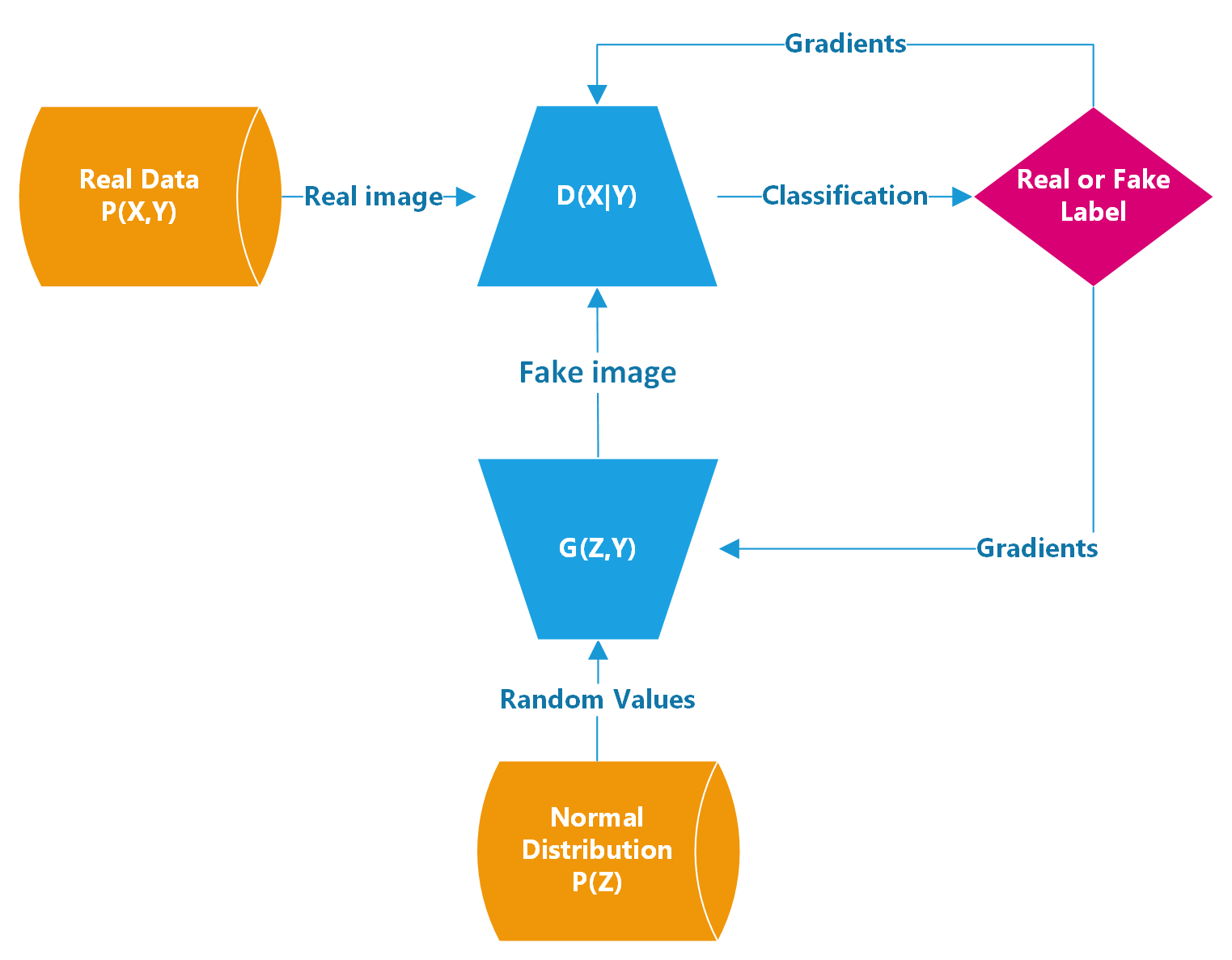}
        \caption{C-GAN Architecture}
        \label{fig:C-GAN-Arch}
    \end{minipage}
\end{figure*}

The Conditional Generative Adversarial Network (C-GAN) is described in Algorithm~\ref{alg:conditional_gan}. The architecture of C-GAN is illustrated in Figure~\ref{fig:C-GAN-Arch}.

For C-GANs, the loss function is based on binary cross-entropy, which is the original GAN loss function proposed by Goodfellow et al. \cite{Goodfellow2014}. The C-GAN loss function aims to optimize the generator (G) and the discriminator (D) through a min-max game. The generator tries to create samples that can fool the discriminator, while the discriminator attempts to accurately classify the samples as real or synthetic. Conditional information, such as class labels, is incorporated into both the generator and discriminator networks. The C-GAN loss function can be expressed as:

\begin{equation}
\begin{split}
L(G,D) = \mathbb{E}_{x,y \sim p_{\text{data}}(x,y)} [\log D(x,y)] \\
+ \mathbb{E}_{z \sim p_z(z), y \sim p_{\text{data}}(y)} [\log(1 - D(G(z,y),y))]
\end{split}
\end{equation}

The architecture of this Conditional Generative Adversarial Network (CGAN) consists of two main components: a generator and a discriminator.

\textbf{Generator:} The generator is responsible for generating new synthetic data. It takes a random noise vector and a class label as input, which are then passed through a series of dense and transposed convolution layers. The output of the generator is a $28\times28$ grayscale image. The generated image is conditioned on the input class label through an embedding layer, a dense layer, and a reshaping operation, after which the label-informed tensor is concatenated with the generator input.

\textbf{Discriminator:} The discriminator's task is to differentiate between real and fake (generated) data. It receives an image and outputs two values: one indicating whether the image is real or fake, and the other indicating the class of the image. The architecture of the discriminator includes convolution layers, batch normalization, LeakyReLU activation functions, dropout layers, and finally, two dense layers for the two outputs.



\textbf{C-GAN Parameters:}
\begin{itemize}
    \item \textbf{Latent Dimension:} The size of the random noise vector that is fed into the generator as an input. This random noise vector, often called a latent vector, is a crucial aspect of Generative Adversarial Networks (GANs). It provides the random seed or the initial point of randomness that the generator will use to produce an output. In this work, we used a latent dimension equal to 100. The size (i.e., the number of elements in the vector) is a hyperparameter and can be tuned for different results.
    \item \textbf{Batch Size:} In this experiment, we used a batch size of 16. In the context of training deep learning models, a 'batch' refers to the subset of the dataset that is used for a single update to the model weights during training. Using a batch size of 16 means that the weights will be updated after 16 examples have been processed. Choosing the right batch size impacts learning as it influences the accuracy of the model's gradient estimate, the stability of the learning process, and the training time.
    \item \textbf{Epochs:} A measure of the number of times all of the training vectors are used once to update the weights in the model. We trained the model for 10 epochs. Running the training process for 10 epochs means going through this entire process 10 times. More epochs could lead to better learning up to a point, after which the model might just be overfitting.
    \item \textbf{Optimizer:} The optimizer is the algorithm used to change the attributes of the neural network such as weights and learning rate in order to reduce the losses. Adam is an optimization algorithm that can be used instead of the classical stochastic gradient descent procedure \cite{amari1993backpropagation} to update network weights iteratively based on training data. The learning rate (0.001) controls how much to change the model in response to the estimated error each time the model weights are updated, and epsilon ($1\times10^{-8}$) is a very small number to prevent any division by zero in the implementation.
    \item \textbf{Loss Functions:} Binary Cross-Entropy loss \cite{ho2019real} is used for binary classification problems. It is suitable when models output probabilities for the two classes. Sparse Categorical Cross-Entropy loss is a form of categorical cross entropy that is very useful when dealing with large categorical output classes. It is used when the labels and predictions are in the form of integers rather than one-hot encodings.
\end{itemize}

In the training loop, the discriminator and the generator are trained in an alternating manner. First, the discriminator is trained on a batch of real samples and a batch of fake samples (generated by the generator). Then, the generator is trained using the combined GAN model, where the discriminator's weights are frozen.

The discriminator tries to correctly classify real and fake images, while the generator tries to generate images that the discriminator cannot distinguish from real images. Over time, both the generator and discriminator improve their capabilities, creating a kind of arms race, leading to the generation of more realistic images.

\subsubsection{C-VAE}
A Conditional Variational Autoencoder (C-VAE) as shown in figure 5 is a variant of the VAE architecture introduced by Kingma et al. \cite{kingma2019introduction}. VAEs are generative models that learn to encode data samples into a lower-dimensional latent space and then decode the latent representations back into the original data space. VAEs consist of two main components: an encoder network that learns the approximate posterior distribution of the latent variables given the data, and a decoder network that learns the likelihood of the data given the latent variables \cite{kim2021conditional}.

C-VAEs integrate conditional information (e.g., class labels) into both the encoder and decoder networks, enabling the generation of data samples conditioned on specific variables. The conditional information is typically concatenated with the input data for the encoder and with the latent variables for the decoder. This modification allows the model to generate more diverse and context-specific samples, which can be beneficial for tasks such as image synthesis, data augmentation, and semi-supervised learning .

\begin{figure*}[!t]
    \centering
    \begin{minipage}{.5\textwidth}
        \centering
        \includegraphics[width=\linewidth]{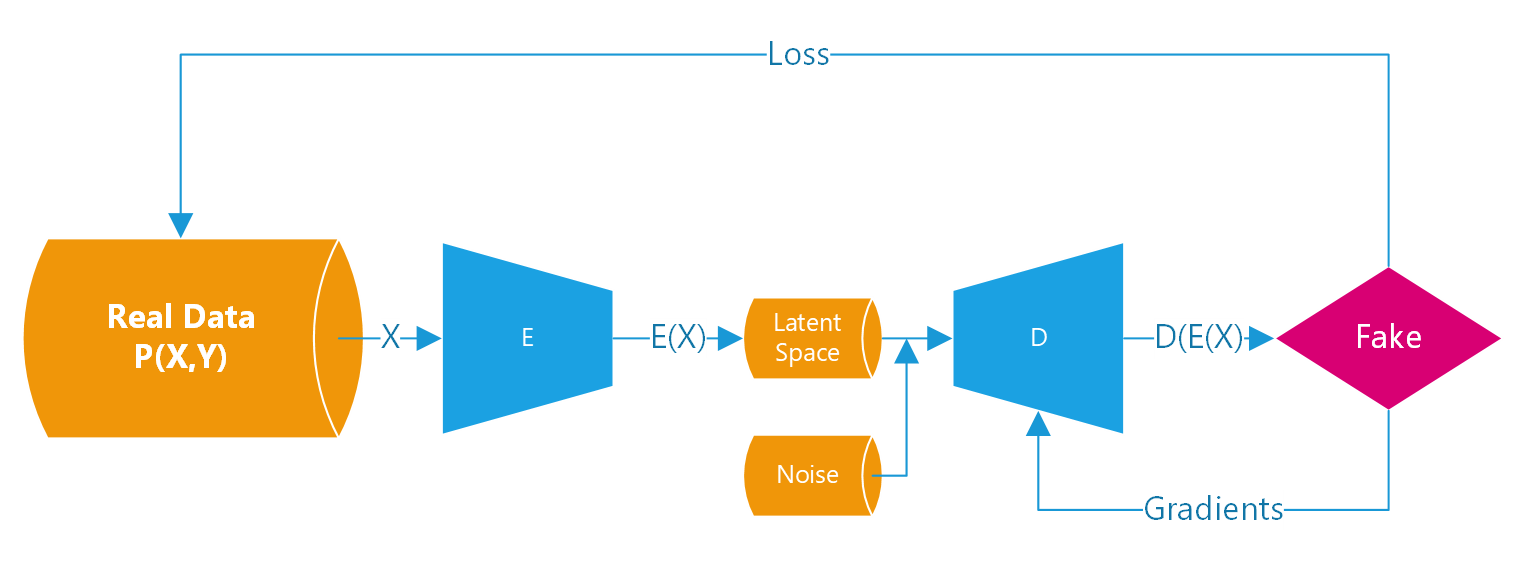}
        \caption{C-VAE Architecture}
        \label{fig:C-VAE-Arch}
    \end{minipage}
    \hfill
    \begin{minipage}{.45\textwidth}
        \begin{algorithm}[H]
            \caption{Conditional VAE}
            \linespread{0.8}\selectfont 
            \label{alg:conditional_vae}
            \begin{algorithmic}[1]
                \State Initialize $E$, $D$, and $Q$ networks with random weights
                \State Set hyperparameters: $\alpha$, $T$, $B$
                \For{$t = 1$ to $T$}
                    \For{each mini-batch $x$, $y$}
                        \State Encode: obtain $\mu$, $\sigma$ from $E(x, y)$
                        \State Sample $z$ from $\mathcal{N}(\mu, \sigma)$
                        \State Decode: generate $\hat{x} = D(z, y)$
                        \State Compute $\mathcal{L}_{\text{recon}} = \sum_i (x_i - \hat{x}_i)^2$
                        \State Compute $\mathcal{L}_{\text{KL}} = -\frac{1}{2} \sum_j (\log(\sigma_j^2) - \sigma_j^2 - \mu_j^2 + 1)$
                        \State Compute total loss: $\mathcal{L} = \mathcal{L}_{\text{recon}} + \mathcal{L}_{\text{KL}}$
                        \State Update $E$, $D$, and $Q$ using gradients from $\mathcal{L}$
                    \EndFor
                \EndFor
            \end{algorithmic}
        \end{algorithm}
        \caption{The Conditional VAE Algorithm}
    \end{minipage}
\end{figure*}

Both C-GANs and C-VAEs have be implemented  using popular machine learning frameworks like TensorFlow \cite{pang2020deep} . This facilitates efficient model development, training, and evaluation, enabling researchers to thoroughly investigate the performance of both generative models in various contexts, including OCR enhancement.
The architecture of the C-VAE is illustrated in Figure~\ref{fig:C-VAE-Arch}.

For C-VAEs, the loss function is a combination of the reconstruction loss and the KL (Kullback-Leibler) divergence \cite{asperti2020balancing}. The reconstruction loss measures the difference between the input data and the generated data after encoding and decoding, ensuring that the C-VAE can accurately reconstruct the input samples. The KL divergence measures the difference between the approximate posterior distribution learned by the encoder and the prior distribution of the latent variables, encouraging the C-VAE to learn a smooth and structured latent space. The C-VAE loss function can be expressed as:

\begin{equation}
\begin{split}
    L_{\text{C-VAE}}(x, y) &= \mathbb{E}_{z \sim q(z|x, y)} [\log p(x|z, y)] \\
    &\quad - D_{\text{KL}}(q(z|x, y) || p(z|y)) \quad
\end{split}
\end{equation}

Both loss functions play a critical role in the training process and determine the effectiveness of the C-GAN and C-VAE models in generating synthetic images for data augmentation and improving the performance of OCR systems.

The optimization algorithm used for both generative models and the OCR model was the Adam optimizer, which has been shown to be effective in training deep neural networks \cite{zhang2018improved}. Training parameters, such as the learning rate, batch size, and the number of training epochs, were selected based on a combination of literature recommendations and empirical testing \cite{bock2018improvement}.

CVAE is structured with an encoder, a decoder, and a reparameterization step in the middle. The encoder converts the input data into a latent representation, while the decoder reconstructs the data back from the latent space.
The \textbf{Encoder} is defined as a class. It's a neural network that transforms input data into two parameters in a latent space, which are means and log\_vars (logarithm of variances). These parameters are used to sample a latent vector, $z$. If the CVAE is conditional (i.e., "conditional" is set to True), the encoder will consider the labels of the input data (encoded as one-hot vectors) along with the input data itself.
The \textbf{Decoder} is also defined as a class. It's another neural network that transforms a given latent vector back to the original data. If the CVAE is conditional, the decoder will consider the labels of the input data (encoded as one-hot vectors) along with the latent vector.
The VAE model is defined to have the encoder and decoder as its components. The forward method of the VAE is where the encoder's output is sampled (using the reparameterization trick to allow backpropagation) and passed through the decoder \cite{johnson2016structured}.

\textbf{C-VAE Parameters:}
\begin{itemize}
    \item \textbf{Number of Epochs:} The C-VAE model is configured with 10 epochs. The term 'epoch' refers to a single iteration where the whole dataset is exposed to the model, encompassing both forward and backward passes. Therefore, in this scenario, the entire dataset will make ten complete journeys through the model.
    \item \textbf{Batch Size:} The batch size has been set to 16. This configuration determines that during the training process, the model will receive sixteen samples from the dataset at every step, again covering both forward and backward passes. While smaller batch sizes may speed up the training process, it could also destabilize it, resulting in more fluctuations in loss. Striking a balance is crucial \cite{sonderby2016ladder}.
    \item \textbf{Learning Rate:} A learning rate of 0.001 is established. The learning rate dictates the extent of alteration in the model parameters in response to the estimated error each time the model's weights are updated. Setting it too high may cause the model to overshoot the optimal solution, while a too low rate may either slow down the training or cause it to stagnate. The value of 0.001, used here, is a common choice for many training scenarios. It is also important to note that the Adam optimizer is being utilized, with a learning rate of 0.001 and epsilon set at $1 \times 10^{-8}$ \cite{sonderby2016ladder}.
    \item \textbf{Encoder and Decoder Layer Sizes:} The architecture of the model is defined by the encoder and decoder layer sizes. The encoder's input layer consists of 784 neurons (28x28, reflecting the size of the input image), followed by a layer with 512 neurons. While this particular architecture is fairly simple with just two layers, more could be added for models that need to handle more complexity. The decoder part of the VAE is also delineated by layer sizes. Its input layer, linked to the latent space, has 512 neurons, followed by a layer with 784 neurons. This reflects a mirrored structure to the encoder but in the reverse sequence \cite{yang2017improved}.
    \item \textbf{Latent Space Dimensionality:} Lastly, the dimensionality of the latent space, otherwise known as the latent size, is set to 10. This parameter determines the size of the compressed representation of the input data. The selection of the latent size represents a trade-off; while a larger latent space might capture more complex representations, it also increases computational complexity and potentially heightens the risk of overfitting.
\end{itemize}

The loss function used for training the CVAE is the sum of the reconstruction loss (Binary Cross Entropy) and the KL divergence, which is a measure of how one probability distribution diverges from a second, expected probability distribution. The Adam optimizer is used to minimize this loss.

\subsubsection{Optimization}
The Adam (Adaptive Moment Estimation) optimizer is a popular optimization algorithm for training deep neural networks, introduced by Kingma et al \cite{kingma2014adam} . It is an extension of the stochastic gradient descent (SGD) algorithm \cite{amari1993backpropagation} that adapts the learning rate for each parameter individually, based on the first and second moments of the gradients. This adaptive learning rate approach allows the optimizer to converge faster and achieve better performance compared to standard SGD. The Adam optimizer computes the first moment (mean) and the second moment (uncentered variance) of the gradients using exponential moving averages. It then corrects the bias in these moment estimates and updates the parameters using the corrected moments. The update rule for each parameter can be expressed as:
\begin{equation}
\begin{split}
    m_t &= \beta_1 m_{t-1} + (1 - \beta_1) g_t, \\
    v_t &= \beta_2 v_{t-1} + (1 - \beta_2) g_t^2, \\
    \hat{m}_t &= \frac{m_t}{1 - \beta_1^t}, \\
    \hat{v}_t &= \frac{v_t}{1 - \beta_2^t}, \\
    \theta_{t+1} &= \theta_t - \alpha \frac{\hat{m}_t}{\sqrt{\hat{v}_t} + \epsilon} \quad
\end{split}
\end{equation}
Here, $g_t$ denotes the gradient at time step $t$, $\beta_1$ and $\beta_2$ are the exponential decay rates for the first and second moment estimates, $m_t$ and $v_t$ represent the first and second moment estimates, $\hat{m}_t$ and $\hat{v}_t$ are the bias-corrected moment estimates, $\alpha$ is the learning rate, $\epsilon$ is a small constant to avoid division by zero, and $\theta_t$ is the parameter at time step $t$. The choice of the Adam optimizer for both the generative models and the OCR model in this study was based on its proven effectiveness in training deep neural networks. Training parameters, such as learning rate, batch size, and the number of training epochs, were selected based on a combination of literature recommendations and empirical testing \cite{sarika2021cnn}.

\subsection{Evaluation}
Fréchet distance, named after Maurice Fréchet, a French mathematician, is a measure of similarity between curves \cite{alt1995computing}. It captures the intuition that two curves are close if you can travel along both curves at the same time while keeping the leash (a line that connects the points on two curves) short. More formally, the Fréchet distance is defined as the minimum leash length that is sufficient for both ends of the leash to traverse their respective curves from start to end. One can imagine walking a dog on a leash: you can move at different speeds, pause, and even reverse, but you can't teleport or leave the path.

In mathematics, the Fréchet distance between two curves in a metric space is a measure of the extent to which the curves are close to each other in their overall shapes rather than merely at individual points. This is particularly useful in many scientific fields, including computational geometry, computer graphics, and GIS systems, where comparing the similarity of different trajectories, paths, or time series data is important \cite{har2014frechet}.

Mathematically, the Fréchet distance can be defined as follows:
Let $C_1: [0,1] \rightarrow M$ and $C_2: [0,1] \rightarrow M$ be two continuous curves in a metric space $(M,d)$, where $d$ denotes the distance in $M$. 
We are looking for reparameterizations $\alpha, \beta: [0,1] \rightarrow [0,1]$, which are continuous and non-decreasing with $\alpha(0) = \beta(0) = 0$ and $\alpha(1) = \beta(1) = 1$. 
The Fréchet distance is then defined as the infimum of all constants $\epsilon \geq 0$ such that there exist reparameterizations $\alpha$ and $\beta$ with the property:
\begin{equation}
\sup_{t \in [0,1]} d(C_1(\alpha(t)), C_2(\beta(t))) \leq \epsilon \quad
\end{equation}
The equation 11 represents the maximum length of the leash needed for a given pair of reparameterizations $\alpha$ and $\beta$.

Therefore, the Fréchet distance is the minimum maximum leash length over all pairs of reparameterizations. In other words, it represents the shortest possible leash that would allow someone traversing the first curve and a dog traversing the second curve to stay connected by the leash for all possible ways of moving along the curves.

In the case of discrete curves (also known as polygonal curves), the computation of the Fréchet distance is significantly simplified because the number of points on the curve is finite. We can imagine each curve as a sequence of points, and we're interested in comparing these sequences in a way that respects the order of the points.

\subsubsection{Fréchet Inception Distance (FID) Score}
The Fréchet Inception Distance (FID) score is a metric used for evaluating the quality of images generated by generative models. It calculates the distance between the real and generated images' feature representations, where these representations are calculated using an Inception model~\cite{szegedy2016rethinking}.

The use of the Inception v3 model is crucial in the FID score calculation because it is capable of extracting high-level features from the images, allowing a meaningful comparison between the real and generated images. It's also important to note that the Inception v3 model is used as a fixed feature extractor and is not trained or fine-tuned during the FID score calculation~\cite{Heusel2017}.

Inception v3 is a convolutional neural network (CNN) developed by Google for image recognition tasks. This model is an enhancement of its predecessors, Inception v1 (also known as GoogLeNet) and Inception v2~\cite{Szegedy2015deeper}.

The defining feature of the Inception v3 architecture is the use of "modules," or small clusters of layers, repeatedly deployed within the network. These "Inception modules" are designed to discern patterns at varying scales within the image.

One such module utilizes factorized $7\times7$ convolutions to reduce computational complexity while preserving the capability to identify high-level features from the input. This is done by breaking down the $7\times7$ convolution into a sequence of $1\times7$ and $7\times1$ convolutions, reducing both the number of parameters and computational cost.

The inception architecture is further extended with the integration of pool projection, another module that includes a pooling operation running parallel to convolution operations. This enhancement increases the model's capacity to handle diverse inputs and combats overfitting.

The architecture begins with a standard convolutional and max pooling layer, succeeded by multiple inception modules. Auxiliary classifiers are included in the network, which helps propagate back gradient information into the deep network, mitigating the vanishing gradients issue.

Generative models can leverage the Inception v3 model to extract features from various layers, enabling the computation of the Fréchet Inception Distance (FID) score. Inception v3, a pre-trained model on the ImageNet dataset, serves as a foundation for feature extraction. The FID score is employed to assess the quality of images generated by these generative models.

Rather than using the default "pool3" layer, corresponding to the 2048-dimensional final average pooling features, other layers can be used:

\begin{itemize}
    \item $64$-dimensional features: These correspond to features extracted post the first max pooling layer. Such features will represent very basic image features such as edges and colors.
    \item $192$-dimensional features: These correspond to features extracted post the second max pooling layer. These features will capture more complex patterns compared to the first layer but will still be relatively low-level.
    \item $768$-dimensional features: These correspond to features extracted before the auxiliary classifier (pre-aux classifier features). These features represent a higher level of abstraction and are closer to the final classification layer.
    \item $2048$-dimensional features: These correspond to features extracted from the final average pooling layer (this is the default in the FID score). These high-level features offer a good balance between low-level image features and high-level semantic content.
\end{itemize}

\begin{figure}[ht]
    \centering
    \includegraphics[width=\columnwidth]{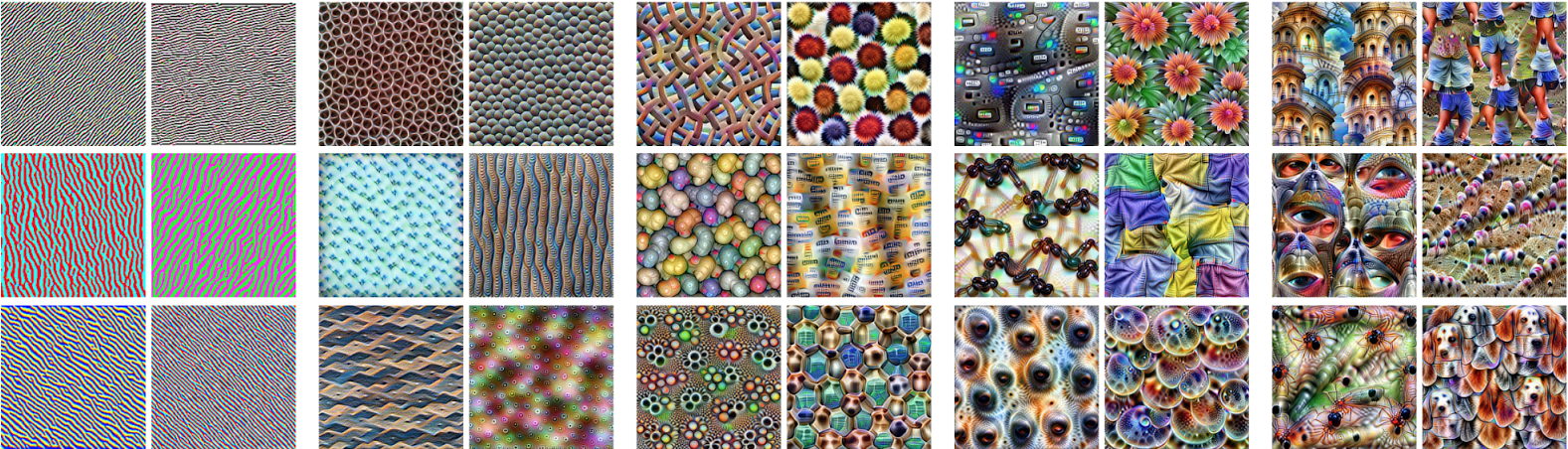}
    \caption{Features learned by Inception v3 trained on the ImageNet dataset~\cite{olah2017feature}}
    \label{fig:inception_features}
\end{figure}

Figure~\ref{fig:inception_features} illustrates the learned features from different level layers of Inception V3, which was trained on the ImageNet dataset. The depicted features span from basic characteristics found in the lower convolutional layers (left) to increasingly abstract features observed in the higher convolutional layers (right).

The model culminates with an average pooling layer and a softmax layer providing the final classification results.

Pre-training a neural network on a large-scale dataset like ImageNet allows the model to learn useful feature representations from the data. When applied to a specific task on a new dataset, it can be fine-tuned, meaning the weights learned during pre-training are minutely adjusted to align with the new data. This approach usually results in superior performance than training a model from scratch, especially when the new dataset is relatively small, as the pre-training assists in preventing overfitting. Thanks to its diverse and high volume of classes, the ImageNet dataset allows models like Inception v3 to learn a broad spectrum of feature representations.

The process of calculating the Fréchet Inception Distance (FID) score starts by preprocessing the real and generated images to align with the input requirements of the Inception model. Typically, this preprocessing involves resizing the images to the input size expected by the model and normalizing the pixel values.

Next, a pre-trained Inception model is loaded. This model is often trained on a large dataset like ImageNet, providing it with a broad understanding of various image features. Once the model is prepared, the real images are passed through the Inception model. The activations from a selected layer are then obtained, capturing the high-level features of the real images.

Similarly, the generated images are also passed through the same Inception model. Activations from the same selected layer are extracted, providing a representation of the high-level features present in the generated images.

The next step involves statistical analysis of these extracted activations. The mean of the activations from the real images, denoted as $\mu_{\text{real}}$, and the mean of the activations from the generated images, denoted as $\mu_{\text{gen}}$, are calculated. Subsequently, the covariance matrix of the activations from the real images, denoted as $\Sigma_{\text{real}}$, and the covariance matrix of the activations from the generated images, denoted as $\Sigma_{\text{gen}}$, are computed.

Further, the squared Euclidean distance between the means of the real and generated activations is computed. This is represented as:

\[ d^2  = |\mu_{\text{real}} - \mu_{\text{gen}}|^2 \]

This measurement offers a quantifiable assessment of the difference between the features of the real and generated images.

Following this, the trace of the sum of the covariance matrices and their square root is calculated. This is expressed as:

\[ \text{trace} = \left(\Sigma_{\text{real}} + \Sigma_{\text{gen}} - 2 \cdot (\Sigma_{\text{real}} \cdot \Sigma_{\text{gen}})^{\frac{1}{2}}\right) \]

This calculation provides an additional measure of the difference in the dispersion of the real and generated image features.

Finally, the FID score is calculated using the following formula:

\[ \text{FID} = d^2  + \text{trace} \]

This score reflects both the difference in the mean and dispersion of the real and generated image features, providing a comprehensive measure of the similarity between the real and generated images. This score is used as a standard measure to evaluate the performance of generative models.

The FID score measures the similarity between the feature distributions of the real and generated images. It takes into account both the differences in means (representing content) and the differences in covariances (representing variety) between the two distributions.

The smaller the FID score, the closer the generated image distribution is to the real image distribution, indicating higher quality and similarity. Conversely, a larger FID score indicates greater dissimilarity between the distributions, implying lower quality and less resemblance to the real images.

It is important to note that FID is generally used with Inception activations from a specific layer of the Inception model, typically the activations from the last layer. The activations represent high-level features extracted from the images, which are then used to compute the mean and covariance. This approach enables a meaningful comparison between the real and generated image distributions.

\subsubsection{FID Score Limits}
While the Fréchet Inception Distance (FID) score is widely used and provides a holistic view of the quality and diversity of generated images, it has its limitations. The traditional FID score calculation relies on high-level features extracted from the Inception model, which may not be suitable for low-fidelity or low-quality images \cite{kohler2012feedback,jung2021internalized,yu2021frechet}.

In our investigation, we discovered that the FID score can be significantly improved by using lower-dimensional Inception activations specifically tailored to low-quality images. By focusing on these lower-dimensional features, such as shapes and edges, which are more relevant and prominent in low-quality images, we can obtain a more accurate measure of image similarity that aligns better with human perception.

However, this realization raises further questions that need to be addressed. We need to investigate the complexity of visual features present in our dataset, in this case, Arabic handwritten digit images, and determine the complexity of features in the pre-trained Inception model that should be used to calculate the FID score. This investigation will enable us to fine-tune the FID computation and enhance the evaluation of generative models in the context of real-time optical character recognition tasks.

By exploring these questions and finding the optimal approach, our research aims to establish a more accurate and efficient method for evaluating generative models when generating images with different feature complexity, specifically focusing on the task of recognizing handwritten character images. The goal is to develop an evaluation method that aligns well with human perception and improves the performance of generative models in real-time optical character recognition applications.

\subsubsection{Low-dimensional Fréchet Inception Distance Score}
We have introduced the Low-dimensional Fréchet Inception Distance (LFID), which is a designed metric that evaluates the discrepancy between the distributions of real and generated images. By focusing on lower-level image features vital for character recognition, LFID allows for a detailed assessment of synthetic image quality in relation to its impact on Optical Character Recognition (OCR) systems. This metric is crucial during training, enabling model alterations to yield images with optimal feature quality, thus improving OCR performance. Furthermore, LFID facilitates early termination of the learning process when models reach desired performance, helping to conserve resources required for model training.
We have adapted the LFID to suit character recognition by lowering feature levels, thus reducing computational load when calculating the Fréchet distance. This modification enhances LFID's suitability for real-time data augmentation, despite its low dimensionality.
Even with reduced dimensions, LFID has been shown to offer accurate image quality evaluations in comparison to the standard Fréchet Inception Distance (FID). By maintaining a robust correlation with human perceptual judgments, LFID guarantees reliable evaluations of generated images.
The LFID allows for the quick evaluation of generated images without sacrificing accuracy, facilitating real-time monitoring and adjustments during training. This immediate feedback helps in spotting and addressing issues with the synthetic images, leading to changes in the model architecture or training parameters. Consequently, this enhances the quality of the generated images, improving OCR performance.
In essence, LFID is an efficient and accurate alternative to traditional FID for monitoring image generation tasks, including OCR. It minimizes the dimensionality of Inception features, speeding up the evaluation of generated image quality without losing accuracy. Hence, it's an ideal choice for real-time monitoring and adjustments during training.
To assess the performance of C-GAN and C-VAE in producing synthetic images and enhancing OCR performance, we utilized a combination of innovative and conventional metrics. The LFID is a crucial part of this evaluation, effectively assessing the quality of images generated by the models during training. LFID's design is computationally efficient while preserving the ability to precisely evaluate image quality.
The study also implemented ablation experiments to assess the significance of different components of the generative models and the OCR systems. This method, which involves the systematic removal or modification of specific components or hyperparameters, enabled the researchers to understand their influence on overall model performance. This broad assessment approach confirmed the findings and shed light on the aspects contributing to the improved OCR performance through generative-based data augmentation.

\subsection{Synthetic Image Evaluation Procedure}
This section on outlines the process for evaluating image quality generated by generative models, primarily using the Low-dimensional Fréchet Inception Distance (LFID) score. We'll cover feature extraction with the Inception V3 model, visualizing learned features, selecting relevant complexity levels, and computing FID score. LFID measures the similarity between real and generated images' feature distributions, guiding model adjustments and hyper-parameter tuning. Early stopping will be discussed as a strategy to prevent overfitting during model training. This section aims to provide a concise understanding of the evaluation process for low-dimensional synthetic images.

\subsubsection{Feature Extraction}
This initial phase of the evaluation process involves extracting meaningful features from images using a large, pre-trained model. The model is the Inception V3 architecture, renowned for its proficiency in handling a wide variety of image-related tasks.

\textbf{Pretrained Model:} We employ the Inception V3 model, which has been pre-trained on the ImageNet dataset~\cite{Krizhevsky2012}. 

Mathematically, the Inception V3 model works by convolving the image with learned kernels, applying batch normalization, and then a rectified linear unit (ReLU) non-linearity, following this general formula:
\[ X_{\text{new}} = \text{ReLU}\left(\text{BN}\left(\text{Conv}\left(X_{\text{old}}, W\right) + b\right)\right) \]
Here, \texttt{Conv} is the convolution operation with kernel $W$, \texttt{BN} stands for batch normalization, \texttt{ReLU} is the rectified linear unit non-linearity, and $b$ is the bias. This operation is performed several times throughout the model to extract and transform features from the image.

Further in the model, these features are flattened and passed through a fully connected layer to generate a fixed-size vector for each image. If we denote the flatten operation as \texttt{Flatten}, and the fully connected layer operation as \texttt{FC}, this can be expressed as:
\[ Z = \text{FC}\left(\text{Flatten}(X_{\text{last}})\right) \]
Here, $X_{\text{last}}$ is the last feature map generated by the convolution operations. The vector $Z$ then represents the extracted features from the image.

The feature extraction process can be viewed as a function $f$ that takes an image $I$ and returns a feature vector $Z$:
\[ Z = f(I) \]
This function $f$ is what we refer to as the Inception V3 model pre-trained on the ImageNet dataset. The vectors $Z$ are what we use in the subsequent stages of the evaluation process.

\subsubsection{Feature Visualization}
After the feature extraction, the next phase involves visualizing the extracted features. This helps in understanding what kind of image properties the model has learned to identify and highlight.

\textbf{Visualizing Learned Features (ImageNet):} We then move on to visualize the features that the Inception V3 model, pre-trained on ImageNet, has learned. This can provide crucial insights into the core patterns and structures the model perceives as significant in an image.

\textbf{Visualizing Learned Features (Fine-Tuned Model):} This involves visualizing the features learned by the Inception V3 model, pre-trained on ImageNet, and subsequently fine-tuned on real images. Fine-tuning often leads to a model better suited to the specific task at hand, in this case, feature extraction. This step allows us to observe how this adaptation process has influenced the model's feature recognition abilities.

\textbf{Selecting Relevant Complexity Level:} Lastly, we identify and select the complexity level of the ImageNet features learned by Inception V3 that are most relevant to our real images, Arabic Handwritten Digit Dataset (AHDD). This provides a focal point to ensure the features maintain the essential characteristics found in the actual images, thereby improving the quality of the synthetic image evaluation. We do this by computing the variance of each feature across the dataset and ranking the features based on this variance. Mathematically, this can be expressed as:
\[ \sigma_i^2 = \text{var}(Z[:,i]), \quad \text{for } i \in \{1, \ldots, n\} \]
Where $Z[:, i]$ represents the $i$-th feature across all images in the dataset. The \texttt{var} function computes the variance, and $\sigma_i^2$ is the variance of the $i$-th feature. n in this context represents the total number of features in the feature vector extracted from the images by the Inception V3 model, with i varying from 1 to n. These features represent the distinct characteristics learned by the model from the images. Inception V3 often is used to extract a 2048-dimensional feature vector from an image when using the final layer before the classification layer, implying that n would typically be 2048 in such a case, but it may vary depending on the specific implementation and layer chosen. Features with higher variance are typically considered more relevant since they capture more variability and, therefore, more information about the dataset.

\subsubsection{Calculating the LFID}
\textbf{Real Images:} Calculate the mean and covariance of the distribution of the lower-level features extracted from the real images. These statistical measures capture the central tendency and dispersion of the feature distribution.

\textbf{Generated Images:} Similarly, compute the mean and covariance of the distribution of the lower-level features extracted from the generated images.

\textbf{LFID:} Use these statistical measures (mean vectors and covariance matrices) from both the real and generated images to compute the Fréchet distance. This distance quantitatively measures the dissimilarity between the two distributions, essentially capturing how far apart the real and generated images are in the feature space.
\[ d^2  = |\mu_{\text{real}} - \mu_{\text{gen}}|^2 \]
\[ \text{Tr} = \left(\Sigma_{\text{real}} + \Sigma_{\text{gen}} - 2 \cdot (\Sigma_{\text{real}} \cdot \Sigma_{\text{gen}})^{1/2}\right) \]
\[ \text{LFID} = d^2  + \text{Tr} \]

\subsubsection{Image Quality Evaluation}
\textbf{Evaluation:} Use the LFID as a measure of the quality of the generated images. A lower LFID indicates that the generated images are more similar to the real images in terms of their lower-level feature distributions, suggesting higher quality.

\textbf{Model Adjustment:}
\textbf{Adjustment:} The need for model adjustment is determined based on the LFID score. From our experiments on the Arabic Handwritten Digit Dataset (AHDD), we have found that an LFID score less than 20 indicates significant similarity between the generated and real images. Therefore, if the LFID score exceeds this threshold, $T = 20$, it suggests that the generated images are not sufficiently similar to the real images, warranting adjustments.  This can be formally expressed as:
\[ \text{If LFID} > T, \text{then adjust model parameters} \]

\textbf{Hyper-parameter tuning:} The adjustments can encompass changes to the model's architecture, modifications to the training parameters, or alterations in the method of image generation. This is typically conducted through an optimization process, such as grid search or Bayesian optimization, which aims to minimize the LFID score.

We search for the optimal values of the parameters used to train the C-GAN and C-VAE. The parameters of interest might include:
\begin{itemize}
    \item The architecture of the CNN, including the number of layers (n\_layers), size of the filters (filter\_size), the size of the stride (stride\_size), batch size (batch\_size), and optimizer (optimizer\_type).
    \item The latent dimension size (latent\_dim).
    \item The batch size (batch\_size).
\end{itemize}

Formally, this can be represented as a minimization problem:
\textit{\begin{align*}
&\text{minimize LFID}(\texttt{n\_layers}, \texttt{filter\_size}, \texttt{stride\_size}, \\
&\texttt{batch\_size}, \texttt{optimizer\_type}, \texttt{latent\_dim})
\end{align*}}

\textbf{Monitoring LFID:} After each adjustment, we monitor the LFID to assess the impact on the image generation quality. Formally, we could express this as:
\textit{\begin{align*}
&\text{LFID}_{\text{new}} = \text{compute\_LFID}(C\text{-GAN or C-VAE with new params}) \\
&\text{If LFID}_{\text{new}} < \text{LFID}_{\text{old}}, \text{then keep the new params}
\end{align*}}

This iterative process of adjustment and monitoring continues until we reach a point where the LFID does not significantly decrease with further tuning, or we hit a preset limit on the number of iterations or time.

\subsubsection{Early Stopping}
The early stopping process is an integral part of model training, designed to halt training when the model's performance starts to plateau or degrade. This process is driven by monitoring the LFID metric during the training of the generative models \cite{yao2007early}.

The LFID score offers a precise assessment of the synthetic image quality, especially in terms of their potential impact on OCR performance. Thus, it is used as a performance criterion to determine when the models have attained optimal performance.
Mathematically, the early stopping condition can be defined as follows:
\textit{\[\begin{gathered}
\text{If } | \text{LFID}_{(i)} - \text{LFID}_{(i-1)} | < \varepsilon, \\
\text{for } i = \text{current epoch}, \text{ then stop training}
\end{gathered}\]}
Here, $\varepsilon$ is a small threshold value that determines the sensitivity of the early stopping procedure. If the absolute change in the LFID score between two consecutive epochs ($i-1$ and $i$) is less than $\varepsilon$, the training is stopped.

\begin{figure}
    \centering
    \includegraphics[width=1\linewidth]{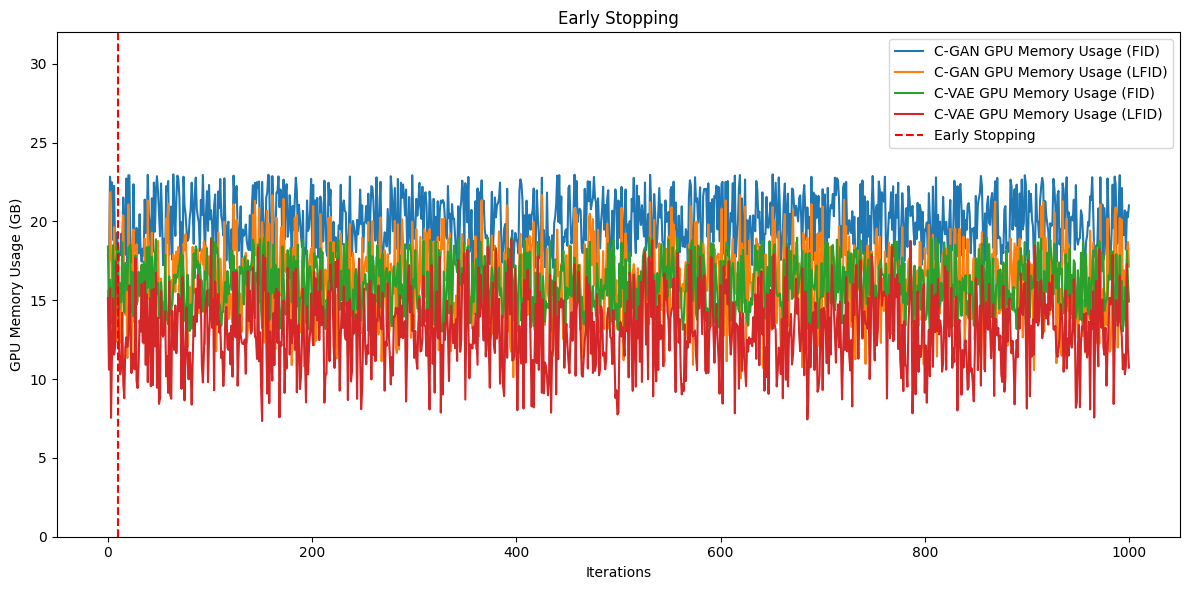}
    \caption{Early Stopping}
    \label{fig:earlysropping}
\end{figure}

This strategy is especially beneficial in two main ways: First, it ensures efficient use of computational resources by avoiding needless training once the model performance has stabilized,as shown in Figure \ref{fig:earlysropping}. 

Second, it helps prevent overfitting, a common pitfall where the model performs exceptionally well on training data but poorly on unseen data.

This mechanism implies a balance between training the model to improve its performance (as measured by the LFID score) and preventing the model from overfitting to the training data, thereby generalizing poorly to new data. The early stopping technique, therefore, plays a critical role in ensuring that the trained model is both efficient and effective.

Please note that the parameter $\varepsilon$ and the number of epochs before triggering early stopping (patience) should be chosen carefully, as too small a value might stop training prematurely, while too large a value might result in overfitting.

Finally, the early stopping technique can be used in conjunction with other regularization techniques and the aforementioned model adjustment process, providing a comprehensive approach to controlling the complexity and improving the performance of the trained models.

\begin{algorithm}[ht]
\setcounter{algorithm}{2}
\caption{Synthetic Image Evaluation Procedure}
\label{alg:3} 
\linespread{0.9}\selectfont 
\begin{algorithmic}[1]
\Procedure {Evaluation}{}

    \State \textbf{Phase 1: Feature Extraction}
    \State $model \gets$ load\_InceptionV3\_pretrained\_model()
    \State $X\_n \gets$ ReLU(BN(Conv($X\_o$, $W$) + $b$))
    \State $Z \gets$ FC(Flatten($X\_l$))
    \State \textbf{Phase 2: Feature Visualization}
    \State Visualize\_features($Z$)
    \State \textbf{Phase 3: Selecting Relevant Complexity Level}
    \For {$i = 1$ to $n$}
        \State $\sigma_i^2 \gets$ var($Z[:,i]$)
    \EndFor
    \State \textbf{Phase 4: Calculating the LFID}
    \State Compute $\mu\_r$, $\mu\_g$, $\Sigma\_r$, $\Sigma\_g$
    \State $d^2 \gets ||\mu\_r - \mu\_g||^2$
    \State $Tr \gets \Sigma\_r + \Sigma\_g - 2 * sqrt(\Sigma\_r * \Sigma\_g)$
    \State $LFID \gets d^2  + Tr$
    \State \textbf{Phase 5: Image Quality Evaluation}
    \State If $LFID > T$ then adjust model parameters
    \State \textbf{Phase 6: Hyper-parameter tuning}
    \State Set $params = (n\_l, f\_s, s\_s, b\_s, o\_t, l\_d)$
    \State Minimize $LFID(params)$
    \State $LFID\_n \gets$ compute\_LFID(C-GAN or C-VAE with new parameters)
    \State If $LFID\_n < LFID\_o$ then keep the new parameters
    \State \textbf{Phase 7: Early Stopping}
    \State If $|LFID_(i) - LFID_(i-1)| < \varepsilon$ then stop training
  
\EndProcedure
\end{algorithmic}
\end{algorithm}

In the context of our Synthetic Image Evaluation Procedure algorithm, we use several abbreviations for brevity and to accommodate space constraints within the algorithm block. The following explains what each abbreviation stands for:

\begin{itemize}
    \item $n\_l$: Stands for the number of layers in the neural network.
    \item $f\_s$: Represents the filter size used in the convolutional layers of the neural network.
    \item $s\_s$: Denotes the stride size, which is the number of pixels shifts over the input matrix.
    \item $b\_s$: Signifies the batch size used in the training process of the neural network.
    \item $o\_t$: Corresponds to the type of optimizer used during the training process.
    \item $l\_d$: Represents the latent dimension in the context of Generative Adversarial Networks (GANs) or Variational Autoencoders (VAEs).
    \item $X\_n$: Stands for the new input tensor after applying the ReLU, BN and Conv operations.
    \item $X\_o$: Denotes the original input tensor before any operation.
    \item $X\_l$: Stands for the last layer's output.
    \item $\mu\_r$: Stands for the mean of the real data.
    \item $\mu\_g$: Signifies the mean of the generated data.
    \item $\Sigma\_r$: Represents the covariance matrix of the real data.
    \item $\Sigma\_g$: Denotes the covariance matrix of the generated data.
    \item $LFID\_n$: Corresponds to the new calculated LFID metric after parameter adjustment.
    \item $LFID\_o$: Represents the old LFID metric before parameter adjustment.
\end{itemize}

We believe that these abbreviations maintain the clarity of the algorithm while efficiently utilizing the available space.

The \textbf{Synthetic Image Evaluation Procedure}(Algorithm \ref{alg:3}) is a multi-phase algorithm for assessing the quality of images generated by generative models. It involves feature extraction using Inception V3, visualization of learned features, selection of relevant complexity levels, and computing the Low-dimensional Fréchet Inception Distance (LFID) score. The LFID score serves as a performance measure to optimize generative models through hyper-parameter tuning and adjustments. Early stopping prevents overfitting. This systematic process improves generative models for low-dimensional synthetic image generation.

\subsection{Saliency Maps}
Convolutional Neural Networks (CNNs) have become a go-to solution for OCR due to their adeptness at handling image data. These networks use a series of filters to extract relevant features from images and then utilize these features to recognize and detect objects, including text characters. However, despite their efficacy, it's often difficult to understand what specific features a CNN uses to detect or classify an object, which might be a problem in critical applications where knowing why a certain prediction was made is as important as the prediction itself. This lack of interpretability or transparency in decision-making is commonly referred to as the 'black box' problem \cite{bilbrey2020look}.

Saliency maps, introduced to tackle this problem, are a class of techniques that highlight the important regions in an image that contribute to the model's final decision, thereby providing a graphical representation of how an AI system interprets visual data. Saliency maps can help us understand where the model is 'looking' when it makes a decision, i.e., which regions of the image it perceives as salient or relevant. They indicate what the CNN model considers critical in an image when identifying an object or a piece of text \cite{adebayo2018sanity}.

Image saliency maps can be particularly effective in OCR systems built on CNNs, as they allow for the examination of the decision-making process of the model. By observing the areas highlighted by the saliency map, researchers and developers can interpret what visual features the model considers crucial for text detection and recognition. This can lead to better understanding of how the model functions, which in turn can be used to improve model performance and reliability, particularly in complex environments where OCR is often put to the test \cite{sharma2012discriminative}.

The use of image saliency maps has the potential to move us a step closer to opening the 'black box' of AI, specifically in the realm of OCR, by offering a visual interpretation of what features are deemed significant by a model. This insight can be instrumental in enhancing not only model transparency but also in designing more efficient and reliable OCR systems in the future \cite{moayeri2022comprehensive}.

In the forthcoming sections of this dissertation, specifically in the "Experiments and Results" segment, we will provide a detailed account of how we utilize saliency maps as an interpretative tool to understand our experimental outcomes. This will encompass the methodology we employed to generate the saliency maps, their practical application in the interpretation of our model's decisions, and the insights we gleaned about the salient features contributing to the model's performance. This process will allow us to provide a comprehensive understanding of our CNN-based OCR model's behavior and reasoning. By visually showcasing what parts of the image our model deemed significant during the detection and recognition process, we will demystify the 'black box' problem and offer greater transparency into the workings of our model, substantiating its performance and reliability.

In conclusion, the methodology chapter has outlined the systematic approach used to evaluate and enhance the image generation capabilities of Conditional Variational Autoencoders (C-VAE) and Conditional Generative Adversarial Networks (C-GAN). With a focus on improving Optical Character Recognition (OCR) systems, our experiments aimed to explore their effectiveness in the domain of Arabic handwritten digit recognition. The Synthetic Image Evaluation Procedure, as described in the preceding section, served as the foundation for conducting our experiments. This comprehensive algorithm allowed us to extract meaningful features, compute the Low-dimensional Fréchet Inception Distance (LFID) score, and optimize the generative models through hyper-parameter tuning and adjustments. In the next section, "Experiments and Results," we will present and discuss the outcomes of these experiments. The results will shed light on the performance and potential of C-VAE and C-GAN for generating high-quality synthetic images, enabling us to gain valuable insights for enhancing OCR systems.

\section{Experiments and Results}

This section is dedicated to describing the series of experiments conducted to analyze and compare the effectiveness of Conditional Variational Autoencoders (C-VAE) and Conditional Generative Adversarial Networks (C-GAN) for image generation in the context of improving Optical Character Recognition (OCR) systems. The emphasis of our experiment was on Arabic handwritten digit recognition \cite{Sawy2017}.
For our experimental approach, we used a pre-established dataset of Arabic handwritten digits. The dataset, exhibiting a wide range of handwriting styles and various degrees of distortion, presents a robust ground for testing the adaptability and efficiency of the OCR systems.

In our initial experiment, we explored the capabilities of both C-VAE and C-GAN in generating diverse synthetic images that bear a resemblance to the original ones in the dataset. The generated images were then introduced into the OCR system to assess its recognition accuracy.

The evaluation of synthetic image quality presents a critical challenge in this domain. To address this, we employed our proposed \textbf{Synthetic Image Evaluation Procedure} (Algorithm \ref{alg:3}) . This evaluation procedure efficiently gauges the quality of the synthetic images produced by the generative models during the training phase. Crucially, it aids in identifying the point of optimal performance, allowing for timely termination of training—a vital aspect considering the usually protracted learning process of generative models.

Furthermore, we adopted Saliency Maps to scrutinize the performance of the upgraded OCR systems. These maps provide insights into which parts of the images the OCR system finds most informative, hence highlighting areas where the system's attention can be optimized.

The experimental results demonstrate that our proposed method of using most realistic synthetic data for enhancing OCR performance offers several advantages. However, it also poses some challenges. Understanding these strengths and limitations can guide future research efforts in this direction and help to devise more precise and efficient OCR systems.

A key highlight of our experiments was the improved performance of the OCR system in recognizing Arabic handwritten digits. Both C-VAE and C-GAN contributed to this enhancement, though they exhibited unique strengths and weaknesses that we discuss in detail. Notably, our proposed evaluation procedure offered faster and more accurate results compared to traditional FID scores, particularly in the context of OCR applications.

As part of this research, we have strived to fortify the performance of Optical Character Recognition (OCR) systems by leveraging data augmentation techniques, primarily through synthetic data generated by generative models. By continuously monitoring and adjusting the balance between the quality and quantity of the images generated, we have aimed to tackle the instability issues that often arise during the training of generative models. The ultimate goal has been to improve the generalization capability of the models, enhance efficiency, and contribute to the optimization of OCR performance.

\subsection{C-GAN}

\textbf{Discriminator:} The input shape is set to $(28, 28, 1)$ for monochromatic images. The class output is designated to have ten classes, signifying the ten Arabic numerals. The Discriminator uses a \textit{Random Normal} distribution for weight initialization, with a mean of $0.0$ and a standard deviation of $0.02$. It is constructed with four convolutional layers, with filters set at $32$, $64$, $128$, and $256$ for each layer respectively. The kernel size is set to $(3, 3)$ with a stride of $(2, 2)$ and 'same' padding to keep the output size consistent. \textit{Batch Normalization} is applied after the 2nd, 3rd, and 4th layers to stabilize and accelerate the learning process. The model utilizes the 'binary crossentropy' loss function for real/fake classification and 'sparse categorical crossentropy' for class label classification. The \textit{Adam} optimizer, with a learning rate of $0.001$ and epsilon of $1e-08$, is used to minimize the loss function.

\textbf{Generator:} The Generator in the C-GAN model takes a $100$-dimensional latent space vector as input. The \textit{Conv2DTranspose} function is used for upsampling with strides of $(2, 2)$. The 'ReLU' activation function is used after Dense layers and the first \textit{Conv2DTranspose} layer, while 'tanh' is used after the final layer to constrain the output values within a suitable range $(-1, 1)$.

For the complete GAN model, similar to the Discriminator, 'binary crossentropy' and 'sparse categorical crossentropy' are used as loss functions for real/fake and class label outputs respectively, and the \textit{Adam} optimizer is again used with the same parameters. This setup ensures the network's learning process is adequately controlled and guided towards generating realistic and distinct images.

\subsubsection{Training Challenges}

The C-GAN model faced significant challenges in generating authentic-looking Arabic Handwritten Digit images, as evidenced in Figure \ref{fig:gan-images}. This issue stems from the multifaceted nature of training Generative Adversarial Networks (GANs), which often necessitates strategic modifications to the model's architecture.

The process requires a delicate balance between the generator's creativity in creating realistic images and the discriminator's ability to discern real from fake. Achieving this balance can be a non-trivial task, and slight deviations can lead to unsatisfactory results. The training challenges were surmounted through iterative experimentation and adjustment of various architectural elements, laying the groundwork for further analysis.

\subsubsection{Training Dynamics}

Training GANs can be likened to a two-player minimax game, involving the simultaneous training of two neural networks: the generator, responsible for creating images, and the discriminator, charged with differentiating between real and generated images.

The delicate equilibrium between these two networks must be maintained, and disruptions in this balance can lead to instability in training. High values of loss functions, particularly in the initial phases of training, underscore this instability, as depicted in Figure \ref{fig:loss}. Achieving convergence in this challenging scenario necessitates careful monitoring and adaptation of training strategies.

\subsubsection{Image Quality and Sharpness}

One of the redeeming features of the C-GAN model is its capability to generate images with pronounced sharpness. Through the adversarial training process, the generator learns to continually refine its output, leading to images with well-defined edges and contours. Figure \ref{fig:gan-images} showcases this ability, illustrating the model's success in producing sharp and visually appealing images despite the noted training difficulties.

\subsubsection{OCR Performance}

\begin{figure}[p]
\centering
\includegraphics[width=0.5\linewidth]{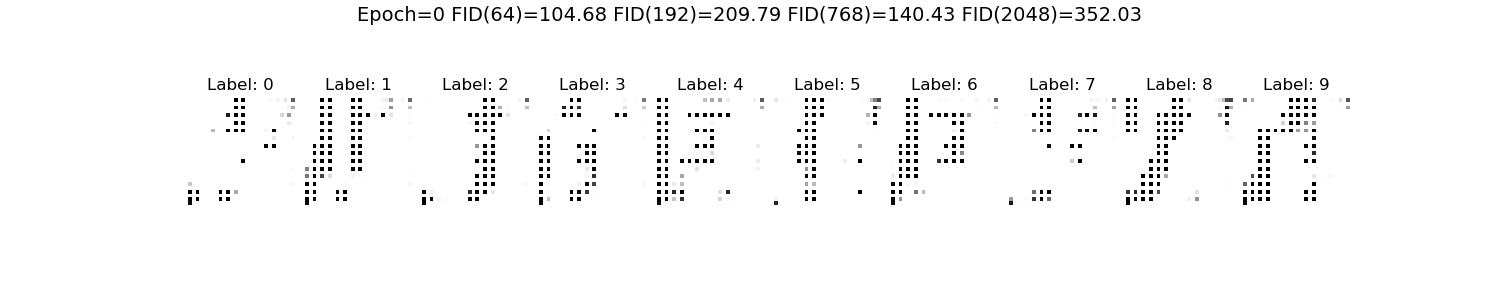}
\includegraphics[width=0.5\linewidth]{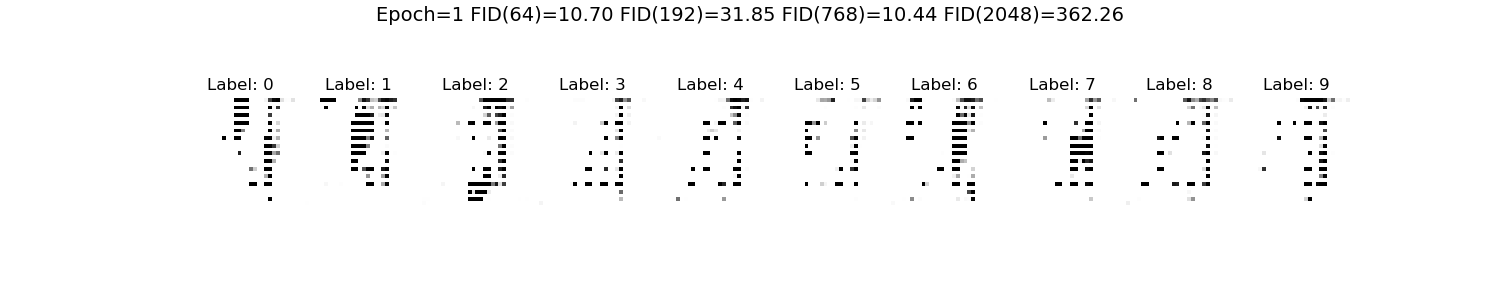}
\includegraphics[width=0.5\linewidth]{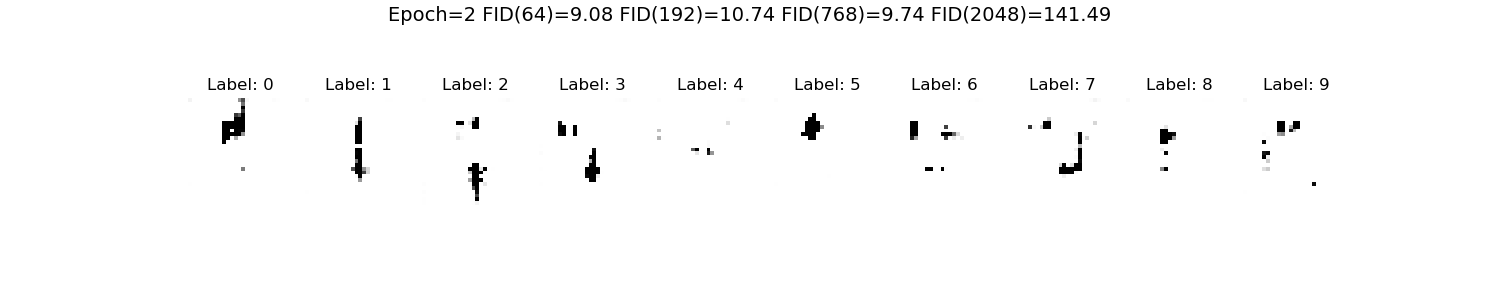}
\includegraphics[width=0.5\linewidth]{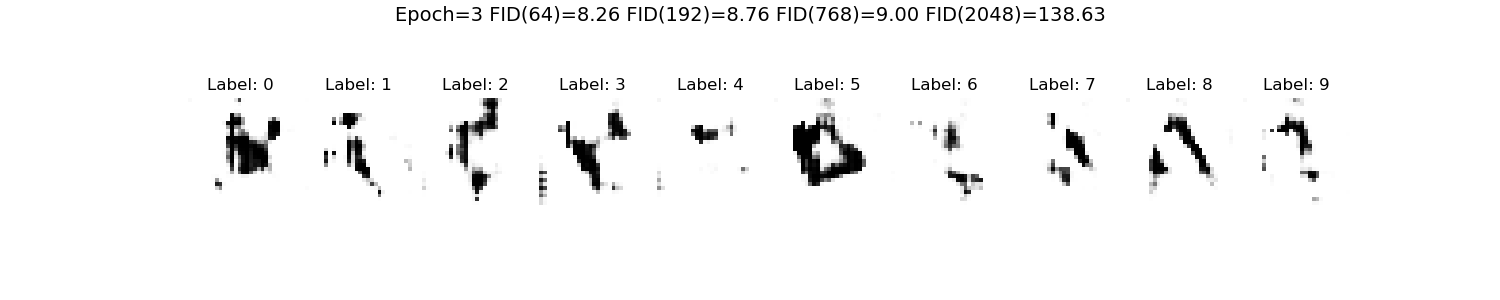}
\includegraphics[width=0.5\linewidth]{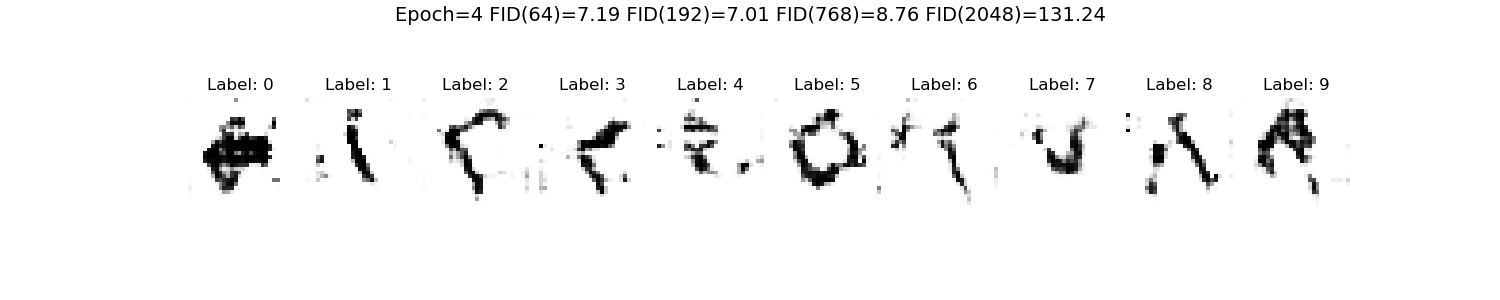}
\includegraphics[width=0.5\linewidth]{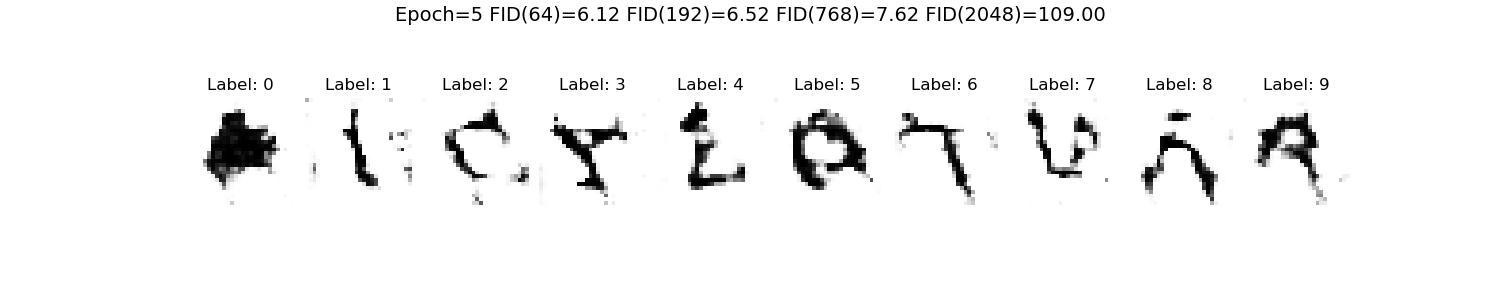}
\includegraphics[width=0.5\linewidth]{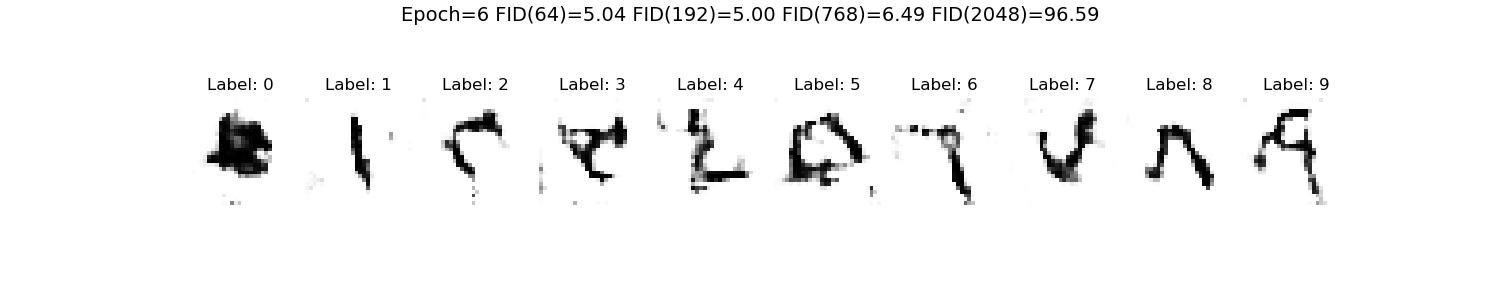}
\includegraphics[width=0.5\linewidth]{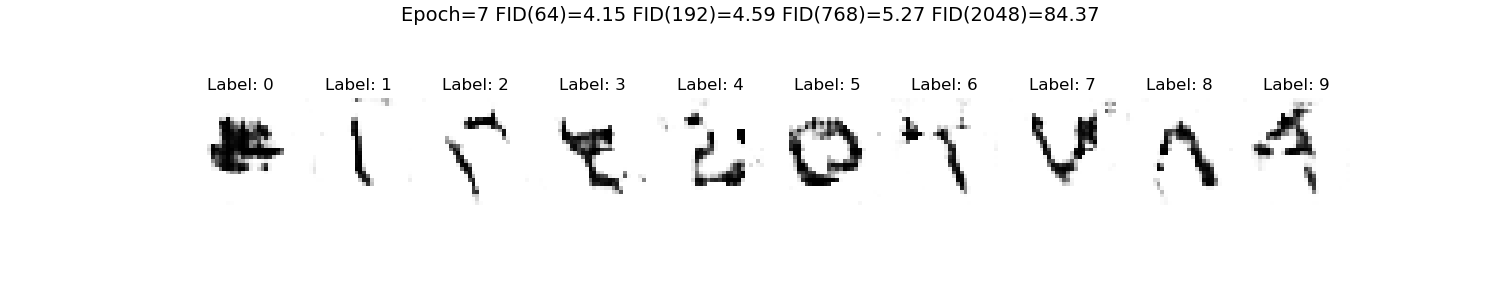}
\includegraphics[width=0.5\linewidth]{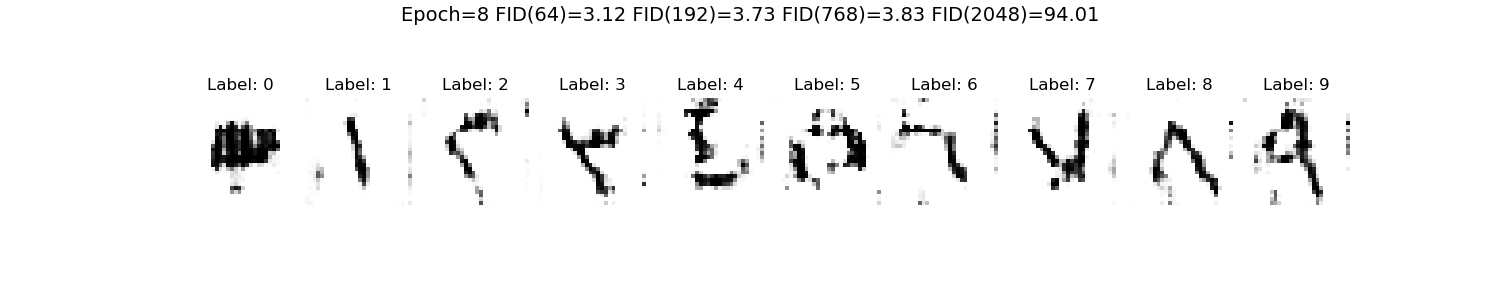}
\includegraphics[width=0.5\linewidth]{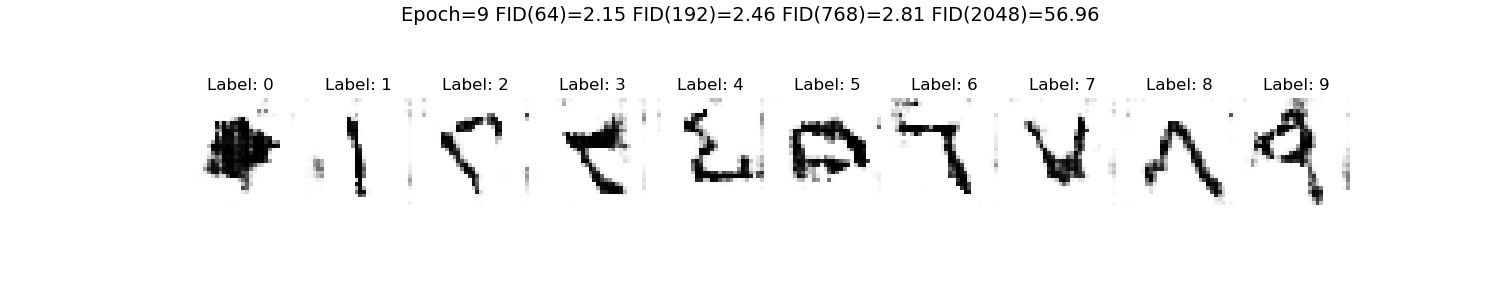}
\caption{Generated Images by the C-GAN Model}
\label{fig:gan-images}
\end{figure}

An intriguing observation was that the enhanced sharpness did not lead to improved performance in OCR systems. Various metrics, such as accuracy, precision, recall, and F1 score, as revealed in Figure \ref{fig:metrics}, highlighted this inconsistency. A closer inspection reveals potential explanations:

Overlooking Crucial Features: The C-GAN might have sacrificed essential features needed for accurate digit recognition in its pursuit of visual appeal. The complexity of representing handwritten digits may have led to this oversight, emphasizing the need for a balanced approach.
Mode Collapse: A phenomenon common in GANs, mode collapse, may have occurred. In this situation, the generator starts producing limited or identical images, failing to capture the diversity in the actual data. This lack of variety could hinder the OCR's ability to generalize, explaining the subpar performance with the C-GAN-augmented dataset shown in Figure \ref{fig:gan-images}.

\subsubsection{Learning Time}

An additional point of interest is the discrepancy in learning time between C-GAN and C-VAE, with the former taking 10 times longer, as seen in Figure \ref{fig:time}. This difference emphasizes the complexities and challenges associated with GAN architecture, reflecting the intricate balance needed for successful GAN training.

\subsubsection{Focus on Unique Features}

The Saliency Maps in Figure \ref{fig:cgan-saliency-maps} highlight that the C-GAN model does not concentrate on unique features within each digit. This lack of focus on critical attributes might contribute to the discrepancies in OCR performance, indicating a potential area for model refinement and further exploration.

\begin{figure}[p]
\centering
\includegraphics[width=0.5\linewidth]{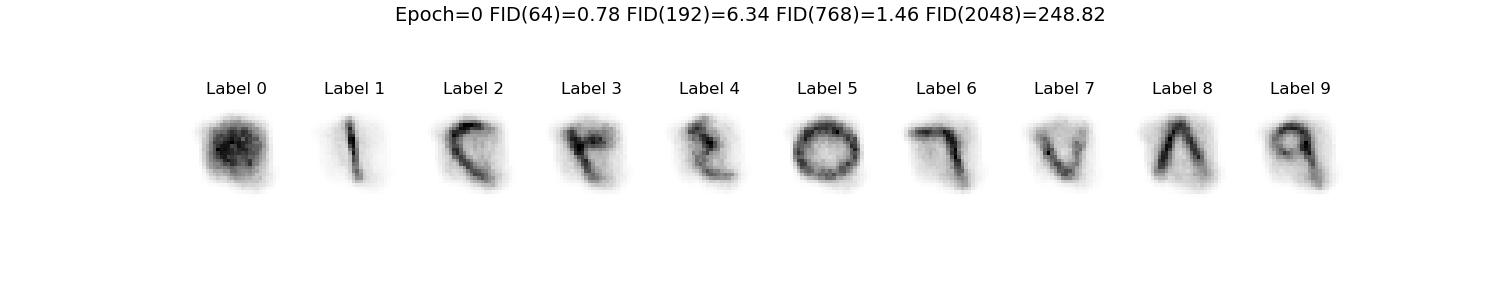}
\includegraphics[width=0.5\linewidth]{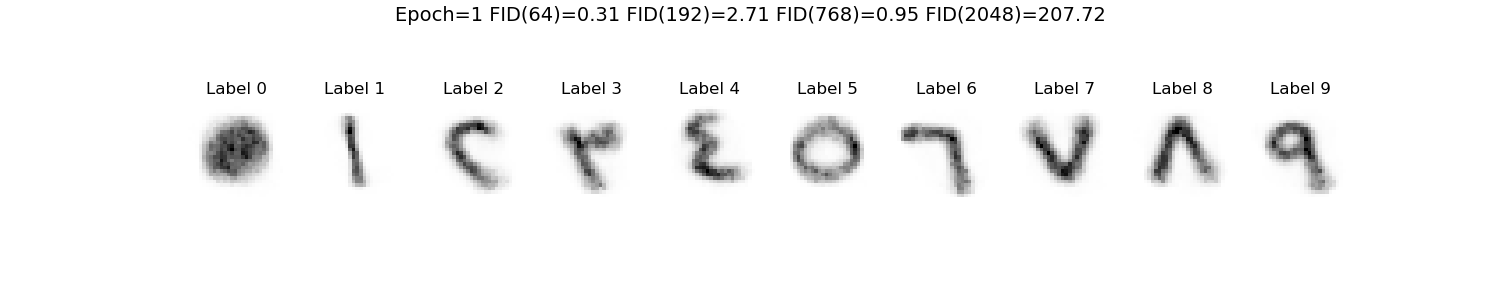}
\includegraphics[width=0.5\linewidth]{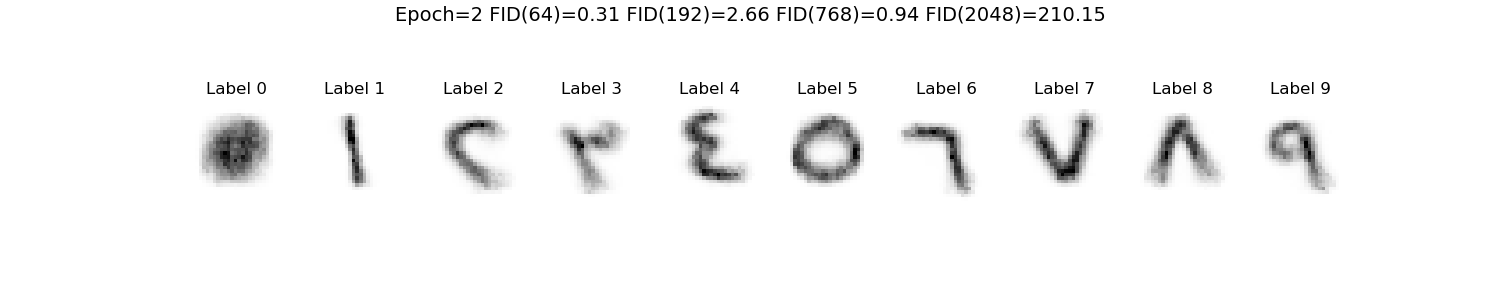}
\includegraphics[width=0.5\linewidth]{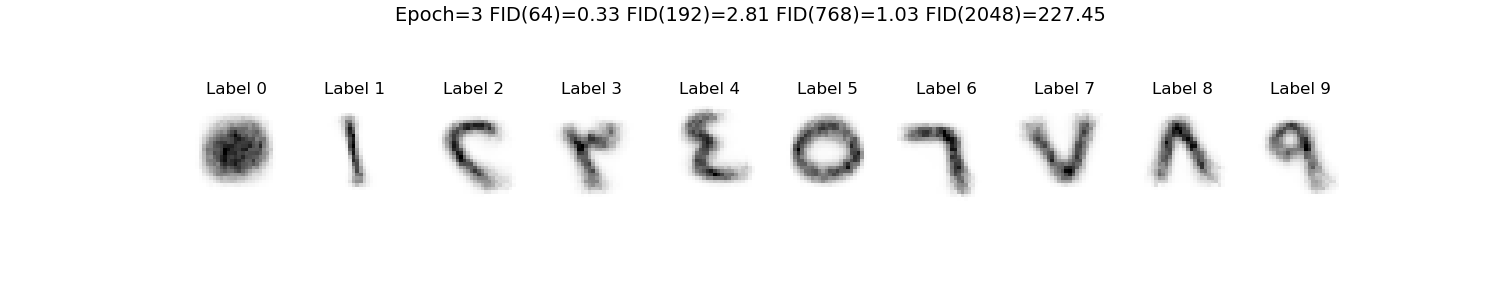}
\includegraphics[width=0.5\linewidth]{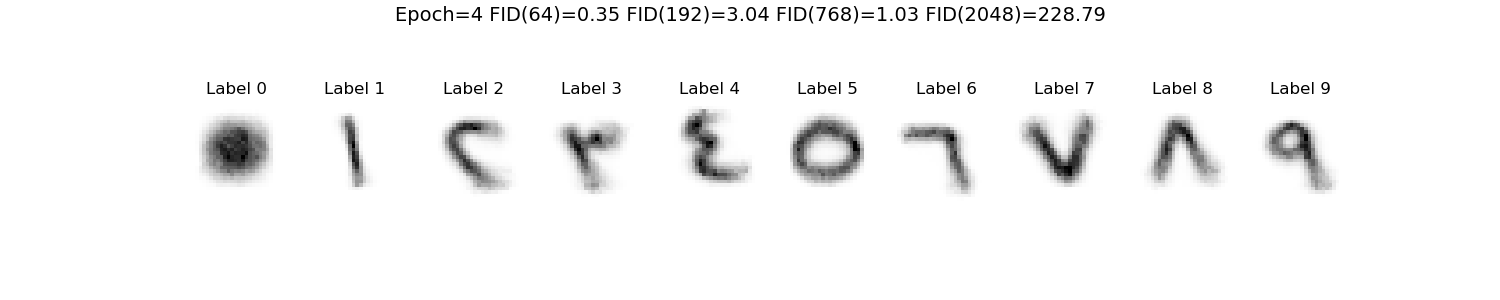}
\includegraphics[width=0.5\linewidth]{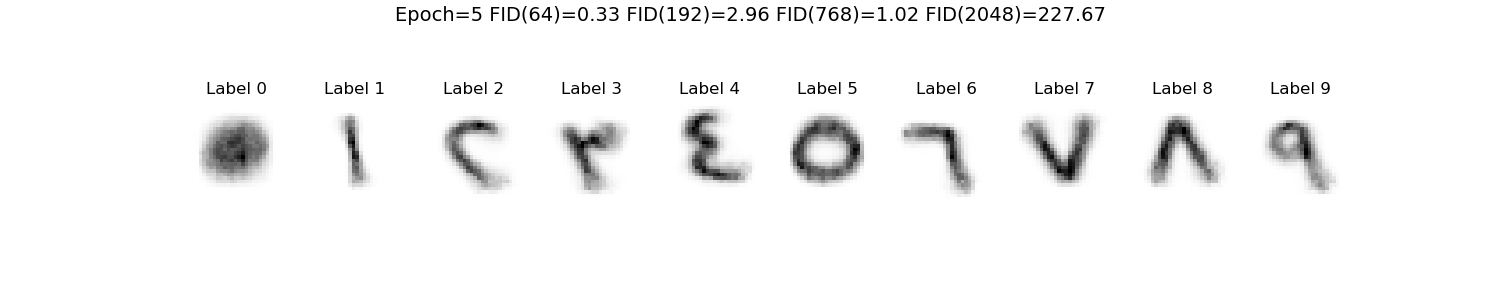}
\includegraphics[width=0.5\linewidth]{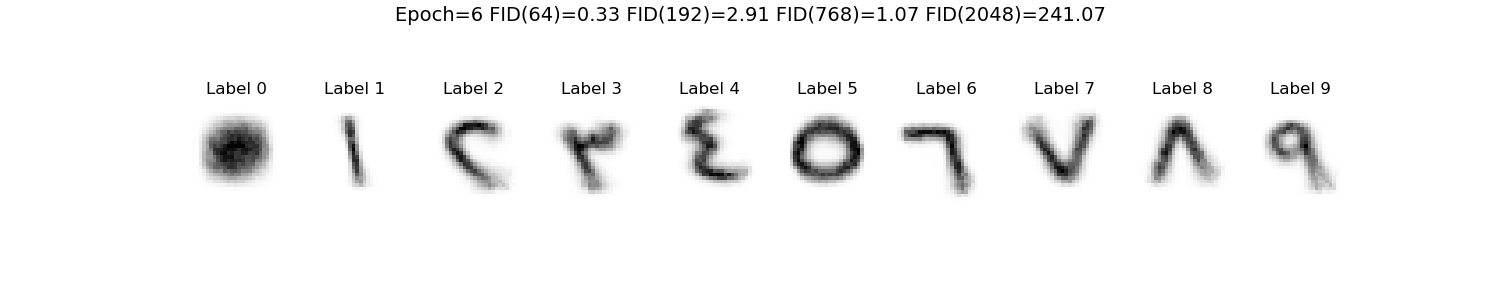}
\includegraphics[width=0.5\linewidth]{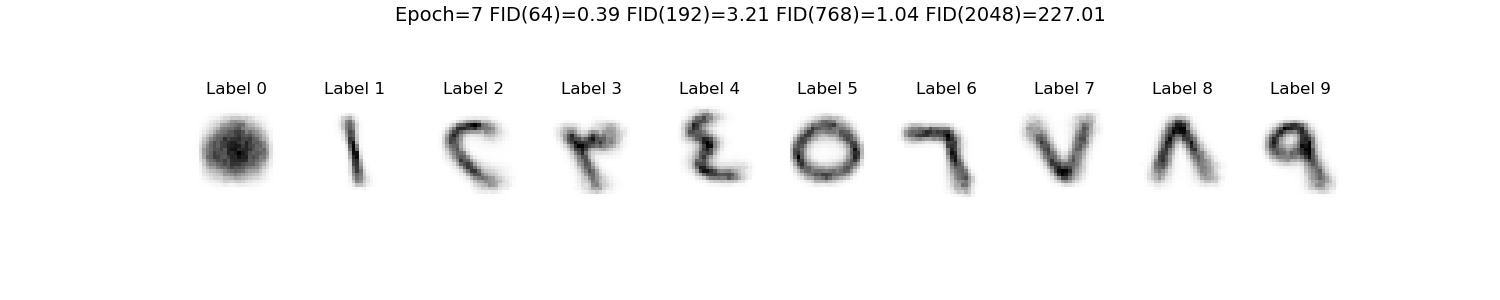}
\includegraphics[width=0.5\linewidth]{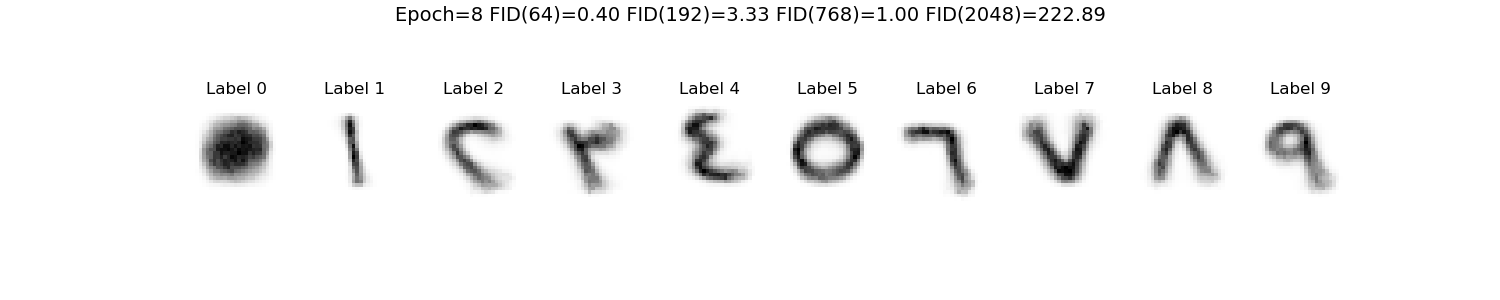}
\includegraphics[width=0.5\linewidth]{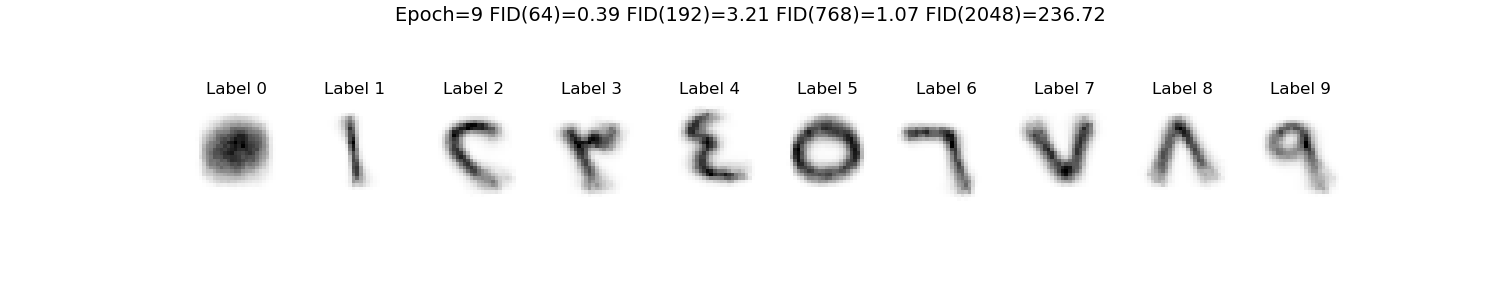}
\caption{Generated Images by the C-VAE Model}
\label{fig:vae-images}
\end{figure}

\subsection{C-VAE}

The Conditional Variational Autoencoder (C-VAE) implemented in this experiment uses several parameters to guide the learning process.

The training process for the C-VAE model is configured to run for 10 epochs. An epoch represents a full pass of the entire dataset through the C-VAE model. The batch size is set to 8, indicating that the model processes eight samples at a time during the training process.

Learning rate, a critical hyperparameter in any optimization algorithm, is set to 0.001. It dictates the size of the steps the model takes towards the minimum of the loss function. A smaller learning rate means the model will learn slowly, but it might result in better performance as it allows the model to fine-tune the weights.

The encoder and decoder architectures of the CVAE model are defined by their layer sizes. The encoder layer sizes are set to [784, 512], which means that the encoder network first maps the input to a 784-dimensional vector, then further down to a 512-dimensional vector. The decoder layer sizes are set to [512, 784], indicating that the decoder network maps the latent representation from a 512-dimensional vector up to a 784-dimensional vector.

The latent size, set to 10, refers to the dimensionality of the space into which the encoder compresses the input data, and from which the decoder generates the output. The loss function for this CVAE model is a combination of Binary Cross Entropy (BCE) and Kullback-Leibler Divergence (KLD). BCE measures the error between the model's output and the actual data, while KLD measures how much the learned latent distribution deviates from a predefined prior distribution.

Adam optimizer is used with a learning rate of 0.001 and epsilon of $1 \times 10^{-8}$ to guide the learning process and ensure optimal convergence of the model's weights. The 'classes' parameter is set to 10, as there are ten classes of Arabic numerals for the model to learn and generate.

\subsubsection{Efficiency in Learning and Image Generation} 
\begin{figure}
    \centering
    \includegraphics[width=1\linewidth]{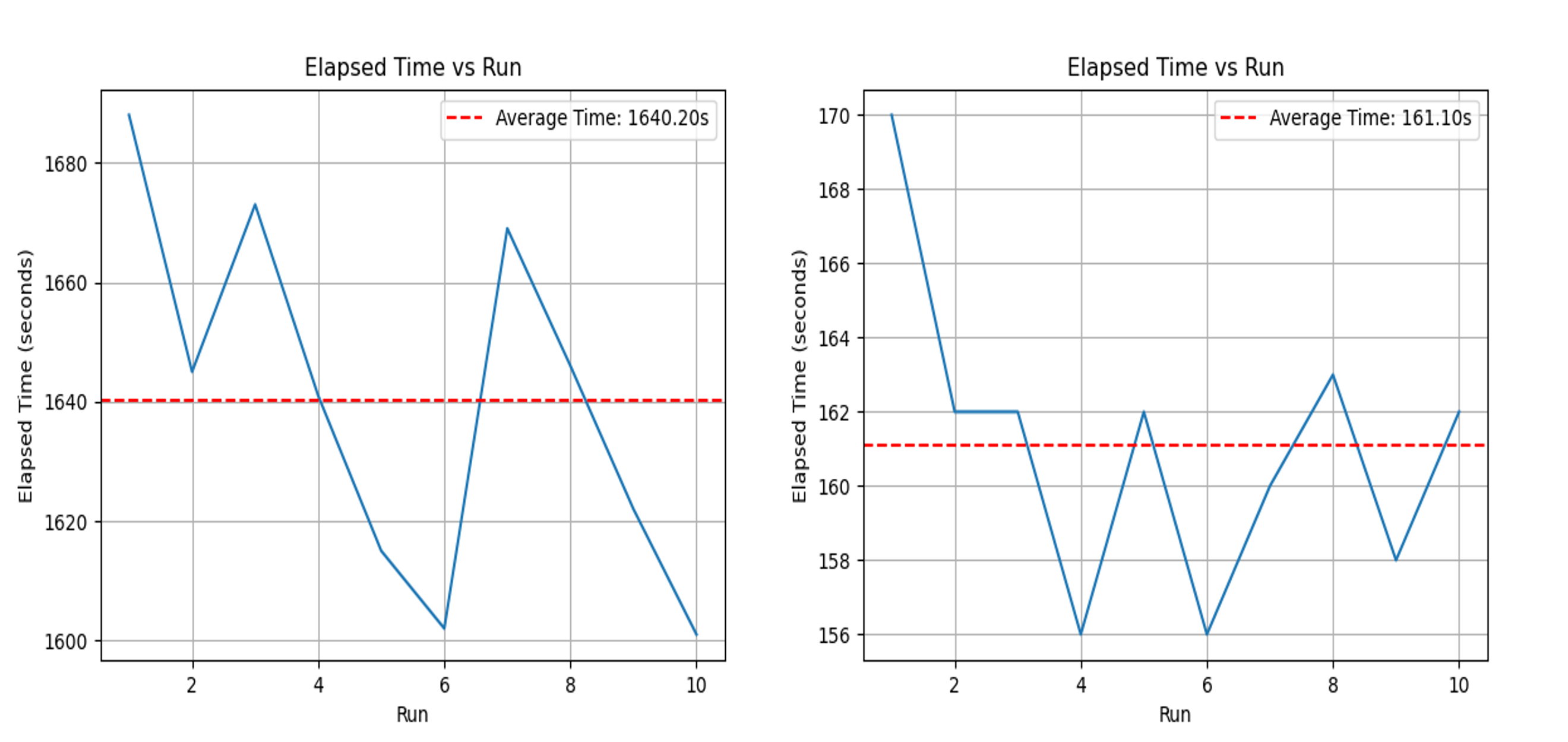}
    \caption{Learning Time (C-GAN on the left and C-VAE on the right)}
    \label{fig:time}
\end{figure}

\begin{figure}
    \centering
    \includegraphics[width=1\linewidth]{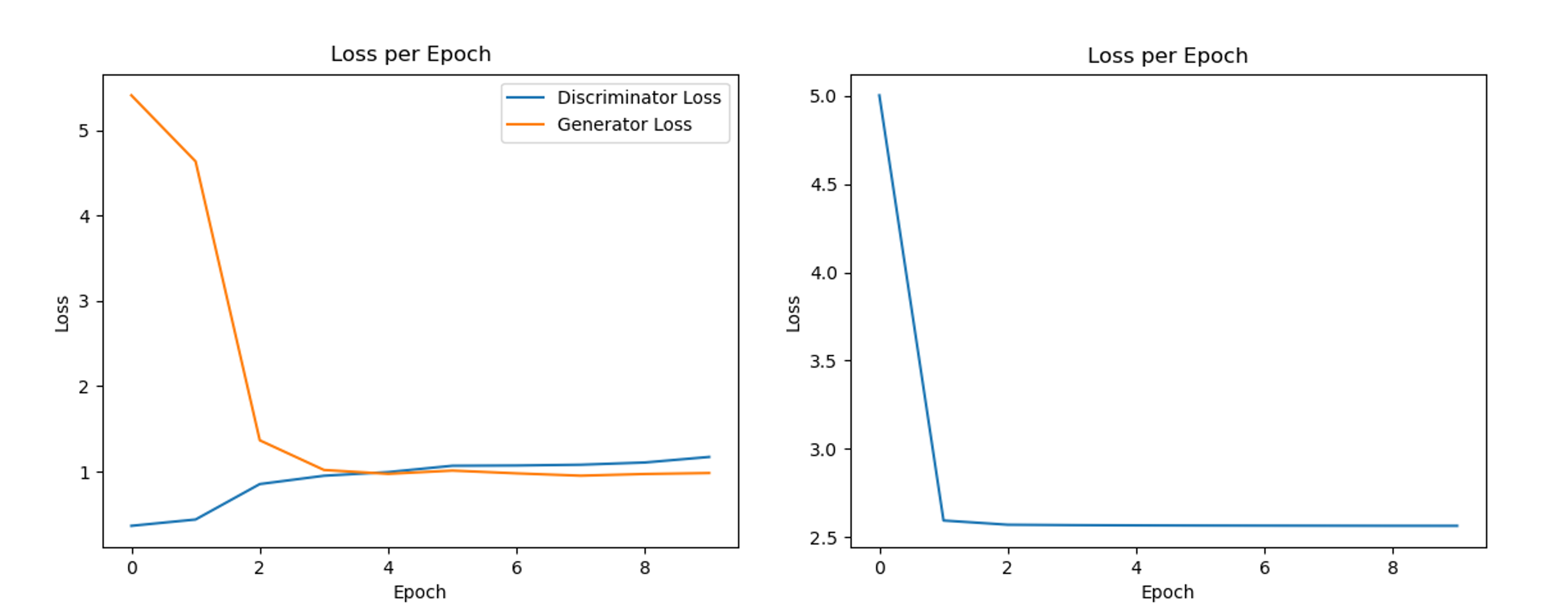}
    \caption{Loss Functions (C-GAN on the left and C-VAE on the right)}
    \label{fig:loss}
\end{figure}

The C-VAE model demonstrated an impressive ability to mimic the Arabic Handwritten Digit dataset swiftly. As seen in Figure \ref{fig:vae-images}, the model began to produce convincing images resembling real handwritten digits after just a single epoch of training. This rapid learning (shown in Figure \ref{fig:time}) progression illustrates the efficiency of C-VAEs in understanding the specific underlying data distribution for this task.

\subsubsection{Blurriness in Generated Images} 
Despite its efficiency, the C-VAE model was not without flaws. A notable shortcoming was the blur in the generated images, a common issue with C-VAEs, as shown in Figure \ref{fig:vae-images}. The origin of this blurriness is traced to the C-VAE's architecture, which promotes a smooth, continuous latent space through the incorporation of a regularization term in the loss function. While this approach ensures continuity, it leads to averaged output over the distribution, resulting in the observed blur.

\subsubsection{Impact on Optical Character Recognition (OCR) Performance} 
Surprisingly, the blurriness in the generated images did not impede the OCR performance. As the metrics in Figure \ref{fig:metrics} demonstrate, the C-VAE-generated images contributed positively to OCR performance. This success can be attributed to the VAE model's ability to retain the essential features necessary for accurate digit classification. Even with the blur, the key attributes distinguishing each digit remained intact, aiding the OCR model in successful recognition.

\subsubsection{Focus on Unique Digit Features:} 
The Saliency Maps presented in Figure \ref{fig:cvae-saliency-maps} further affirmed the C-VAE model's focus on unique features within each digit. By concentrating on these characteristics, the model ensured that the crucial elements for digit classification were preserved.

\subsection{Training Procedure}
The training procedure for both C-GAN and C-VAE models consists of several crucial steps, designed to effectively train the models and optimize their performance in generating synthetic images for OCR tasks. These steps include:

\begin{enumerate}[label=\arabic*.]
    \item Mini-batch Preparation: The Arabic Handwritten Digits training dataset is preprocessed, and mini-batches of 64 real images and their corresponding class labels are created.
    \item Noise and Label Sampling: For each mini-batch, random noise vectors are sampled for the C-GAN model, and latent vectors are sampled from the encoder's output distribution for the C-VAE model. These vectors are combined with class labels.
    \item Generator/Encoder Training: The C-GAN generator is trained to generate synthetic images from noise vectors and class labels, while the C-VAE encoder is trained to encode input images into a latent space representation while considering class labels.
    \item Discriminator/Decoder Training: The C-GAN discriminator is trained to distinguish between real and generated images while considering class labels, and the VAE decoder is trained to reconstruct input images from sampled latent vectors and class labels.
    \item Latent Space Regularization: An essential component of C-VAE training is the regularization of the latent space using the KL divergence loss. This loss encourages the model to learn a smooth and meaningful latent space representation.
    \item Training Iterations: The training process involves updating the generator/encoder and discriminator/decoder weights to improve their performance progressively. Limiting the number of training iterations helps balance performance and computational efficiency, preventing overfitting.
\end{enumerate}

\subsection{Monitoring Progress and Loss Functions}
During the training process of GAN and VAE models, it is crucial to keep track of progress and loss functions to guarantee convergence, as well as to identify and address potential issues. This monitoring helps to optimize the models' performance and provides valuable insights into the training dynamics.

\begin{enumerate}[label=\arabic*.]
    \item \textbf{LFID Metric}: Periodically calculating the LFID metric allows for the assessment of the quality of the generated images. This computationally efficient metric provides real-time feedback on the performance of the models, which can be used to fine-tune the training process and make necessary adjustments to improve image generation quality.
    \item \textbf{Model Losses: }Tracking different loss components for each model type is essential for understanding how well the models are learning and adapting during the training process.
    \begin{itemize}
        \item \textbf{C-VAE} it's important to track both the reconstruction loss and the KL divergence loss. The reconstruction loss measures the difference between the input images and their reconstructions, while the KL divergence loss enforces a smooth and meaningful latent space representation by ensuring it follows a specified prior distribution, typically a standard Gaussian distribution.
        \item \textbf{C-GAN} monitoring the generator and discriminator losses is vital. The generator loss quantifies how well the generator is able to create realistic images, and the discriminator loss measures how accurately the discriminator can distinguish between real and generated images.
    \end{itemize}
\end{enumerate}

By consistently monitoring progress and loss functions, the training process can be optimized, ensuring that the models converge and achieve the desired performance in generating synthetic images for OCR tasks. This monitoring process enables the identification of potential issues and allows for adjustments to the models or training parameters to improve overall performance.

\subsection{Early Stopping}
Incorporating early stopping criteria based on the LFID metric or validation dataset performance can be employed to prevent overfitting and reduce training time if the models converge faster than anticipated. This strategy helps optimize the models' performance by stopping the training process when further training would not lead to significant improvements, ensuring efficient use of computational resources.

In summary, the detailed training procedure for GAN and VAE models focuses on optimizing their performance in generating synthetic images for OCR tasks. By incorporating mini-batch preparation, noise and label sampling, generator/encoder and discriminator/decoder training, latent space regularization, training iterations, monitoring progress, and employing early stopping criteria, the training process aims to effectively train the models, prevent overfitting, and achieve the desired balance between performance and computational efficiency.

\subsection{OCR}
A Convolutional Neural Network (CNN) is employed as the OCR model for Arabic handwritten digit recognition. The architecture of the CNN consists of two convolutional layers, each followed by a max-pooling layer, a dropout layer, and a fully connected layer for classification. The CNN is designed to learn and extract important features from the input images, facilitating accurate digit recognition.

Training the OCR Model: To investigate the effects of data augmentation on the performance of the OCR model, the CNN is trained separately on the original dataset and each augmented dataset. The training process involves using mini-batch training with a suitable batch size and training the model for a predefined number of epochs. This allows for a comparison of the OCR model's performance when trained on different datasets, providing insights into the impact of the C-GAN and C-VAE-generated synthetic images on the model's accuracy and generalization capabilities.

By augmenting the dataset with synthetic images and training the OCR model on the original and augmented datasets separately, the study aims to assess the effectiveness of different data augmentation strategies and their impact on the performance of the OCR model. This approach helps to understand the potential benefits of using GAN and VAE-generated images for OCR tasks and identify the most effective augmentation strategy for improving model performance.

\subsection{Evaluation}
\begin{figure}[ht]
    \centering
    \includegraphics[width=1\linewidth]{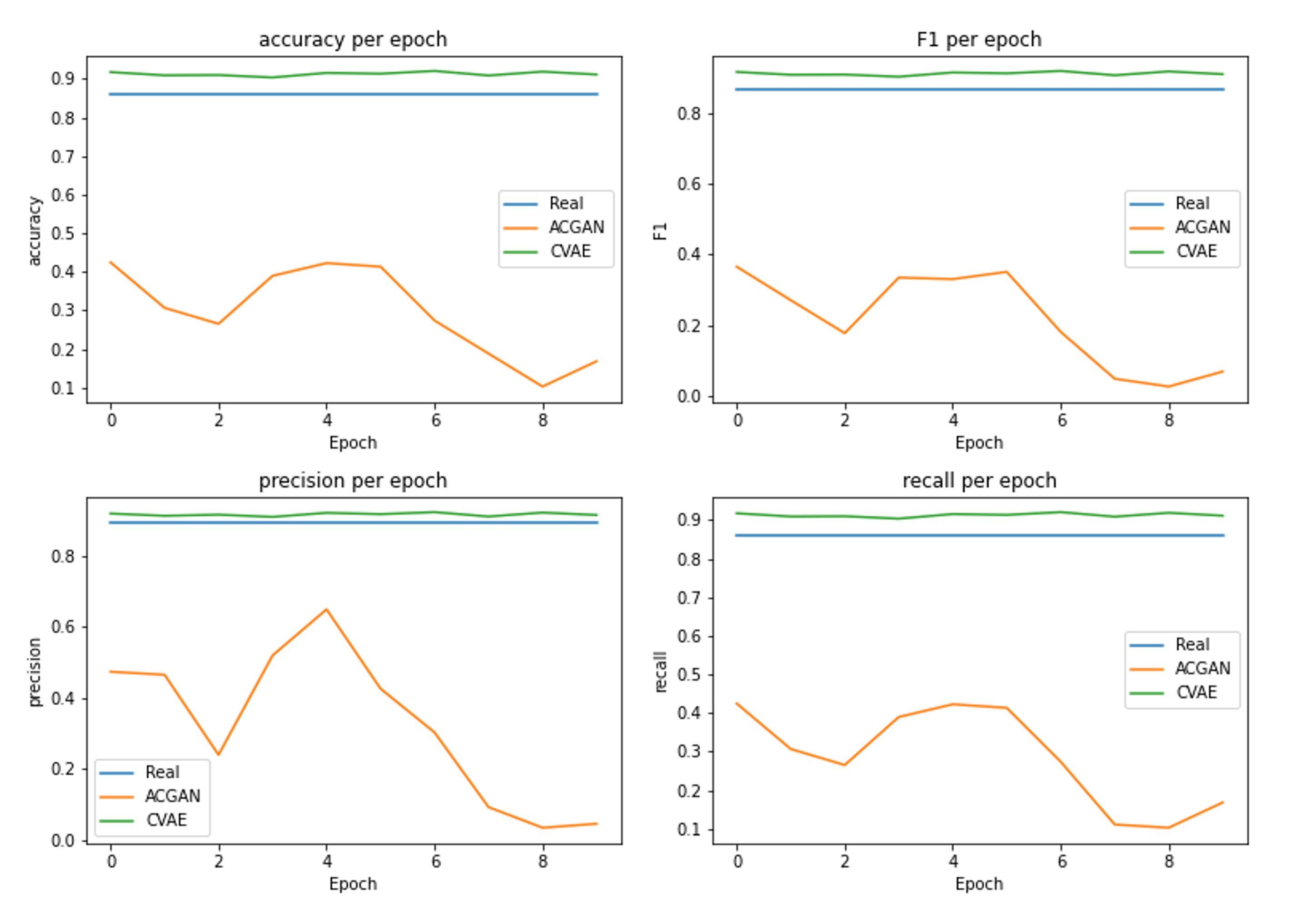}
    \caption{Classification Metrics}
    \label{fig:metrics}
\end{figure}

Evaluating the OCR Model: After training the OCR model on the original and augmented datasets, its performance is evaluated on a separate test dataset to assess generalization capabilities and compare the impact of different data augmentation strategies. Accuracy is used as the primary evaluation metric, but other metrics like precision, recall, and F1-score can also be reported to provide a more comprehensive assessment of the model's performance, as shown in the Figure \ref{fig:metrics}.

Investigating the Impact of Synthetic Data: To understand the relationship between the quality of synthetic images, as measured by the LFID metric, and the improvement in OCR performance, the correlation between LFID scores and OCR model accuracy on the test dataset is investigated. A strong correlation would indicate that the LFID metric is effective in evaluating the quality of generated images and that higher quality images lead to better OCR performance.

Comparing OCR Features with Arabic Handwritten Digit Features: Another way to assess the impact of synthetic data on OCR performance is to compare the most critical features learned by the OCR model with the unique features of Arabic handwritten digits. This analysis can help identify whether the synthetic images generated by GAN and VAE models capture essential features needed for accurate digit classification. It also provides insights into the effectiveness of different generation loss functions and data augmentation strategies in preserving and enhancing these features for better OCR performance.

By evaluating the OCR model on test data and investigating the correlation between LFID scores and model accuracy, as well as comparing important OCR features with the unique features of Arabic handwritten digits, the study aims to provide a comprehensive understanding of the impact of synthetic data on OCR performance. This approach helps to identify the most effective augmentation strategies and generation loss functions for improving model performance in OCR tasks.

\subsection{FID Scores}
The Fréchet Inception Distance (FID) score is a metric used to evaluate the quality of images generated by generative models. In essence, a lower FID score suggests that the distribution of features extracted from the generated images is closer to the distribution of features extracted from real images, indicating better quality. This study confirms that a lower dimensional FID (LFID) score correlates positively with the perceived quality of the images generated by the Conditional Variational Autoencoder (C-VAE) as compared to the Conditional Generative Adversarial Network (C-GAN).

Two primary pieces of evidence support this claim. First, human perception of the similarity between generated and real images aligns with the LFID scores. Generated images from the C-VAE model, which have lower LFID scores, are perceived as more similar to the real images than those generated by the C-GAN model with higher LFID scores. This perception of likeness suggests that the C-VAE model is more successful at creating images that closely mimic the properties of the real images.

Second, the comparison of LFID scores provides quantitative support to this observation. The C-VAE model's lower LFID score signifies a smaller statistical distance between the real and generated images, indicating that it has successfully learned a more accurate representation of the original data distribution. Conversely, the C-GAN model's higher LFID score suggests a greater distance, indicating a less accurate representation. Thus, the LFID scores not only substantiate the human perception of image quality but also provide a quantifiable measure of the quality and realism of the images generated by these models.

Adding to the argument of the effectiveness of lower dimensional Fréchet Inception Distance (LFID) over the traditional high-dimensional FID scores, our experimental results bring out another crucial aspect. It was observed that images generated through the Conditional Variational Autoencoder (C-VAE) significantly improved the performance of the Optical Character Recognition (OCR) system, as opposed to those generated by the Conditional Generative Adversarial Network (C-GAN), which, interestingly, had an adverse effect on OCR performance.

The generated images from the C-VAE model, associated with lower LFID scores, contributed to enhancing the OCR's ability to recognize and interpret the characters accurately. The generated images effectively enriched the training dataset and facilitated the OCR model in learning a more robust and generalized representation of the digit classes.

\begin{figure}[ht]
    \centering
    \includegraphics[width=1\linewidth]{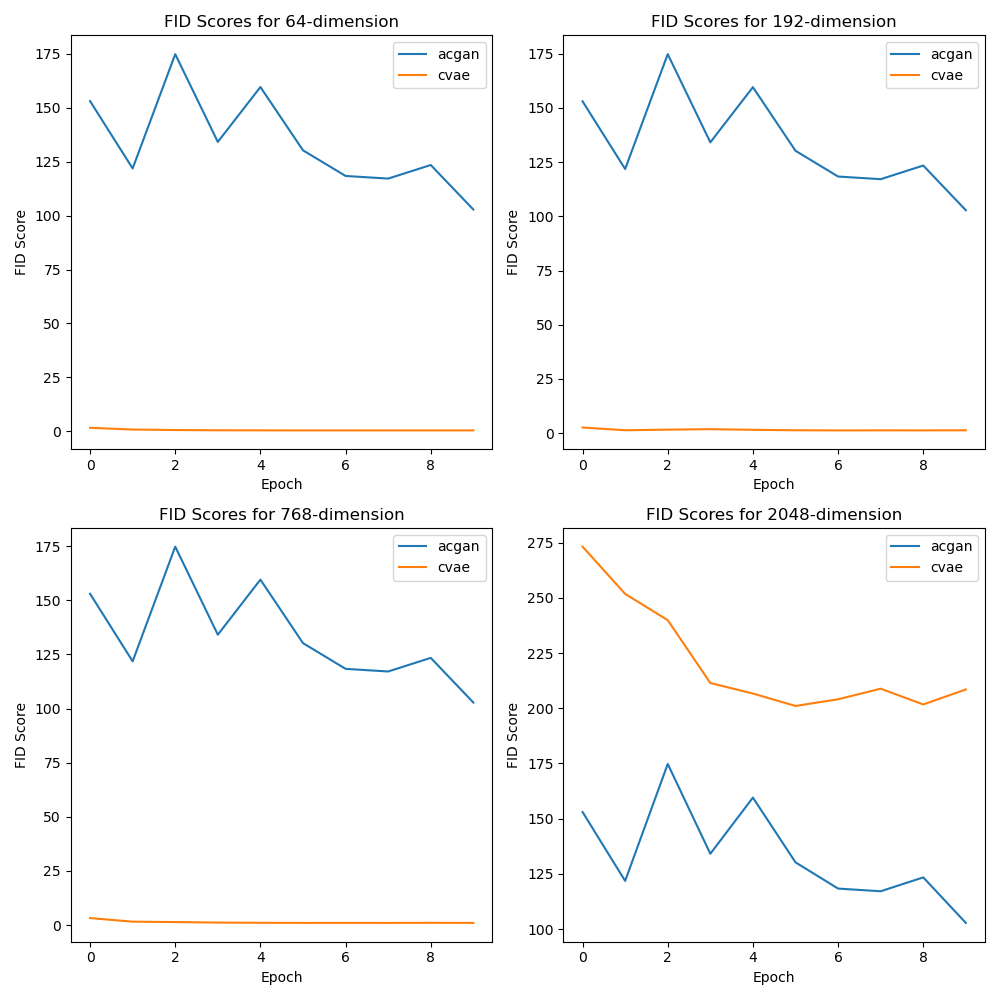}
    \caption{FID Scores}
    \label{fig:fid}
\end{figure}

Contrastingly, C-GAN, despite producing sharper images, couldn't contribute positively to OCR performance. C-GAN generated images, which were associated with higher LFID scores, seemed to compromise the OCR's interpretive capacity. This inverse relationship between the LFID scores and the OCR performance reaffirms the credibility of LFID as a more meaningful and insightful measure, especially in lower dimensions, in evaluating the quality and utility of images generated by these generative models.

Therefore, our experimental results add another dimension to the advantages of LFID over high-dimensional FID, emphasizing its effectiveness not just in reflecting the perceptual similarity and the statistical likeness between the generated and real images, but also in predicting the utility of the generated images in improving the performance of downstream tasks, like OCR in this case.

\subsection{Saliency Maps Visualization }
\begin{figure}[ht]
    \centering
    \includegraphics[width=0.22\linewidth]{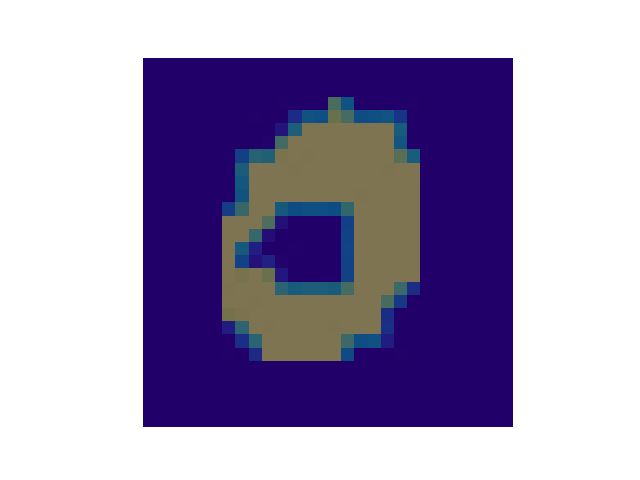}
    \includegraphics[width=0.22\linewidth]{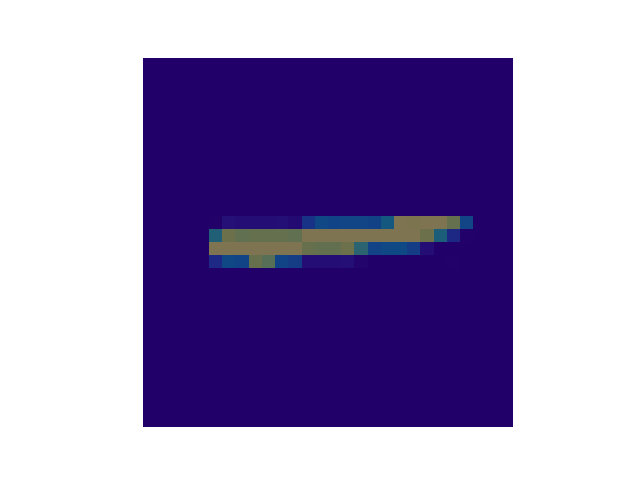}
    \includegraphics[width=0.22\linewidth]{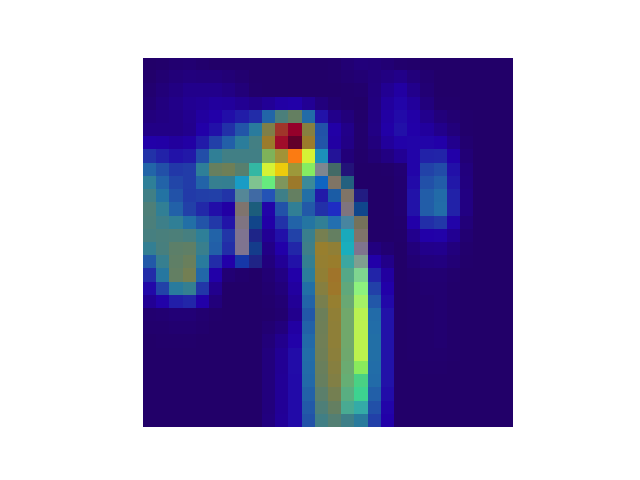}
    \includegraphics[width=0.22\linewidth]{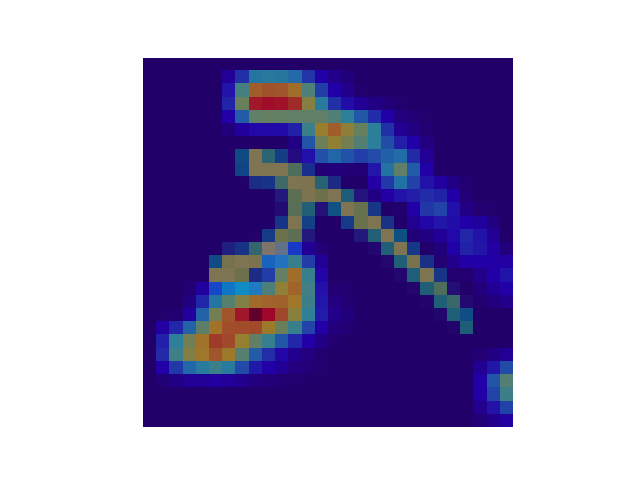}
    \includegraphics[width=0.22\linewidth]{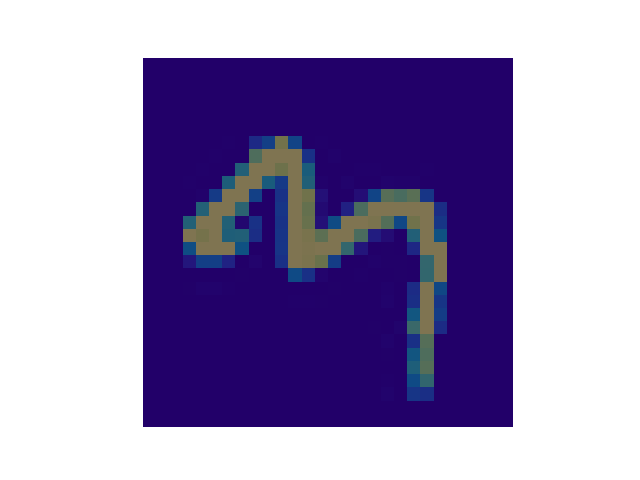}
    \includegraphics[width=0.22\linewidth]{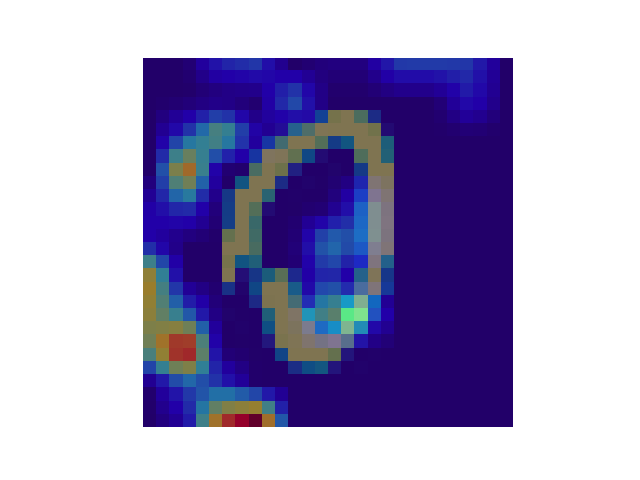}
    \includegraphics[width=0.22\linewidth]{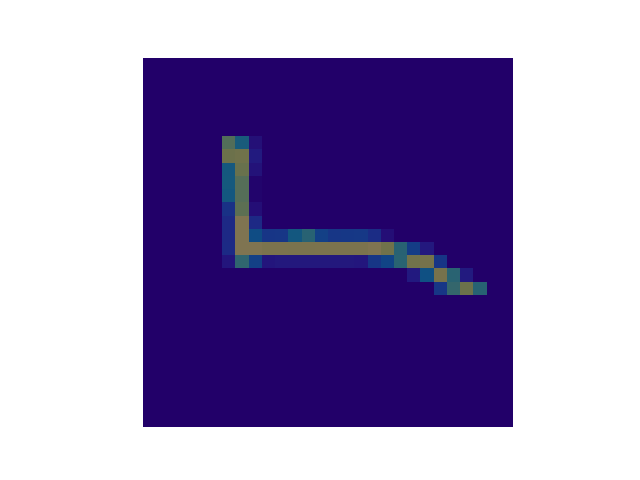}
    \includegraphics[width=0.22\linewidth]{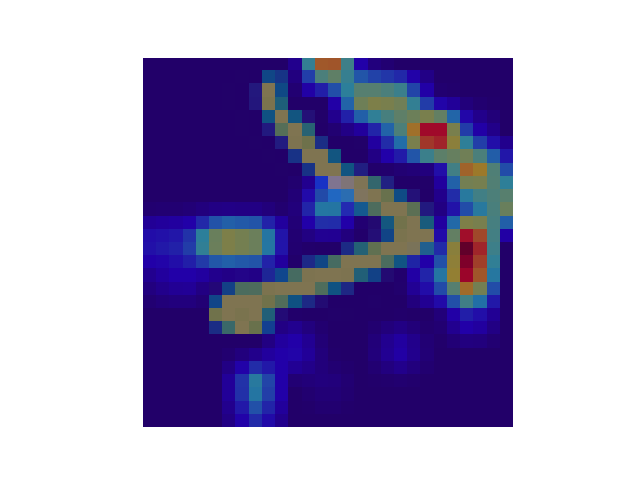}
    \includegraphics[width=0.22\linewidth]{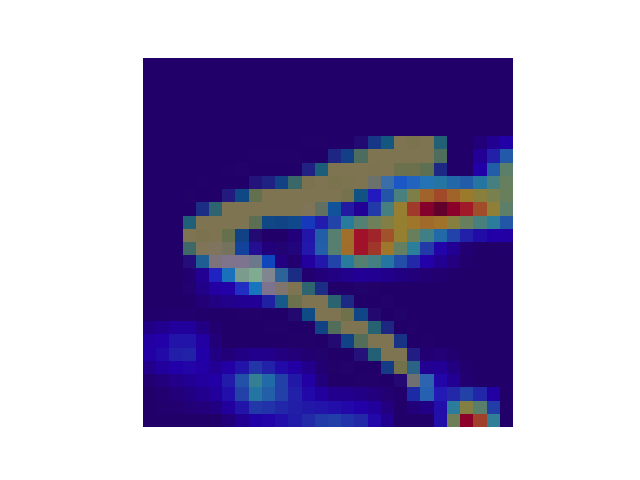}
    \includegraphics[width=0.22\linewidth]{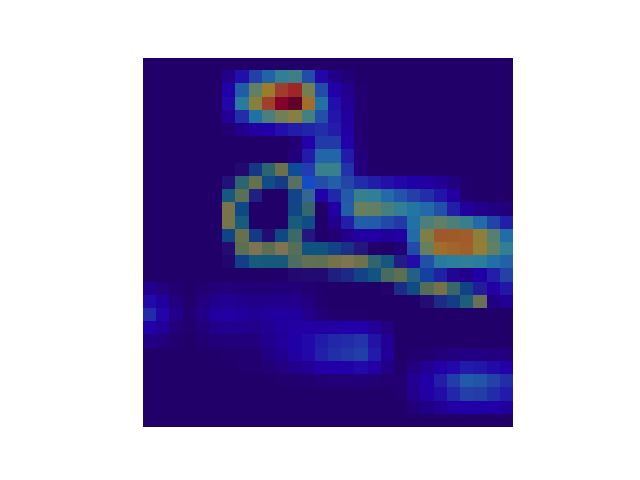}
    \caption{C-GAN Saliency Maps}
    \label{fig:cgan-saliency-maps}
\end{figure}

\begin{figure}[ht]
    \centering
    \includegraphics[width=0.22\linewidth]{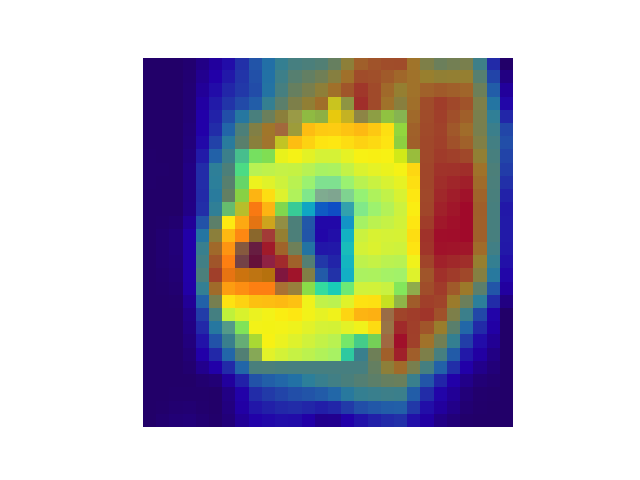}
    \includegraphics[width=0.22\linewidth]{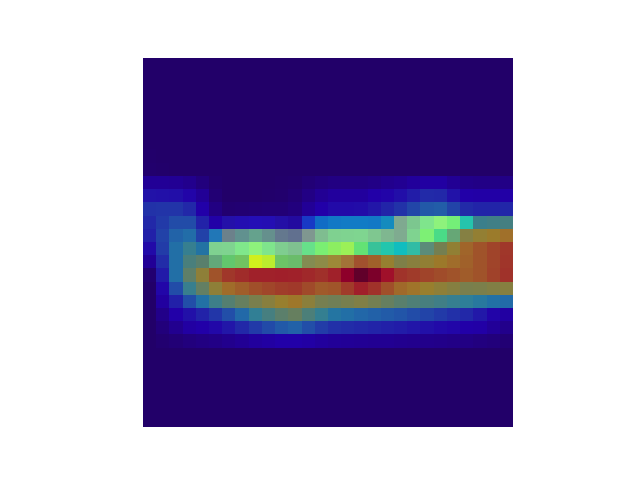}
    \includegraphics[width=0.22\linewidth]{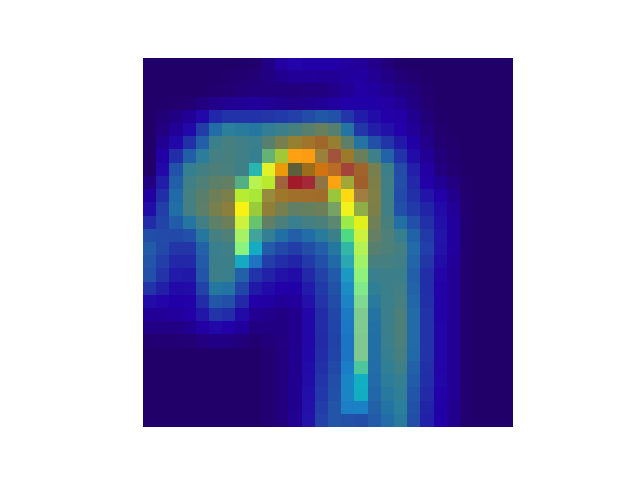}
    \includegraphics[width=0.22\linewidth]{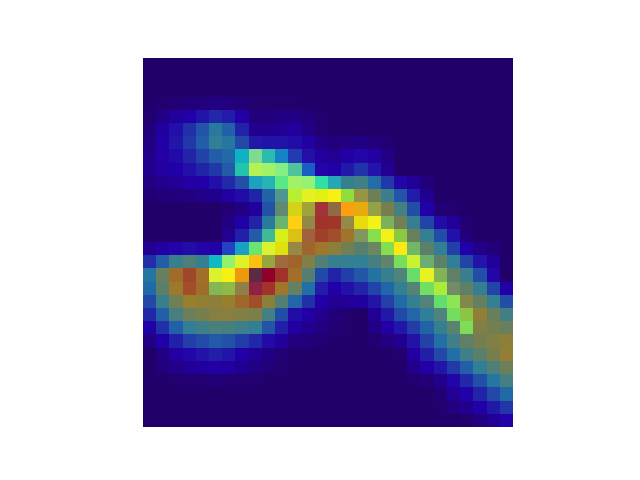}
    \includegraphics[width=0.22\linewidth]{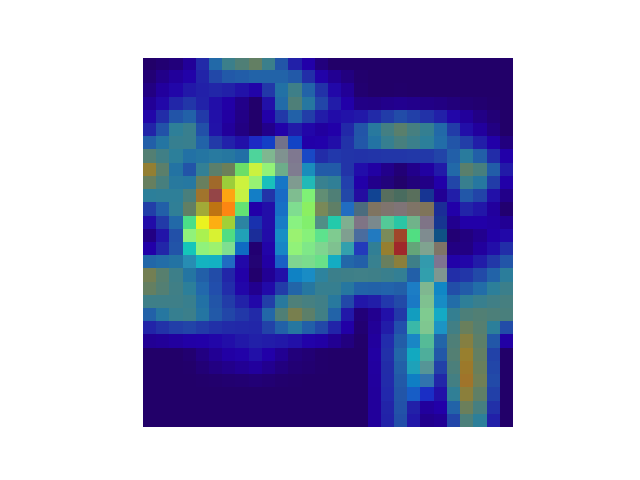}
    \includegraphics[width=0.22\linewidth]{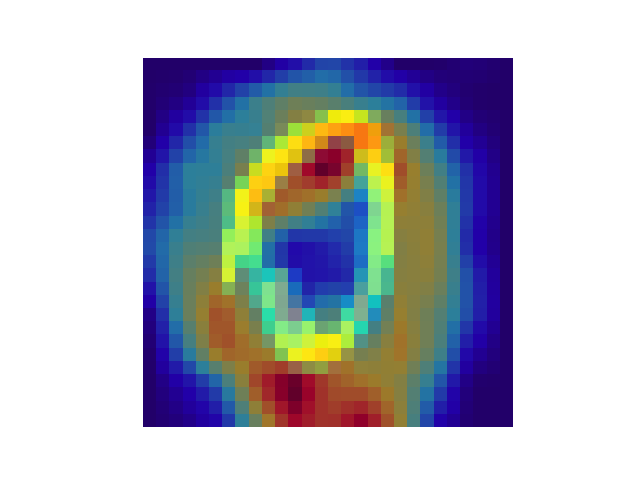}
    \includegraphics[width=0.22\linewidth]{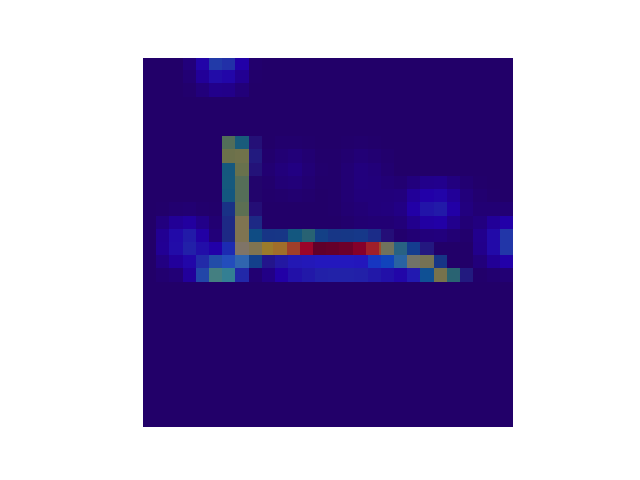}
    \includegraphics[width=0.22\linewidth]{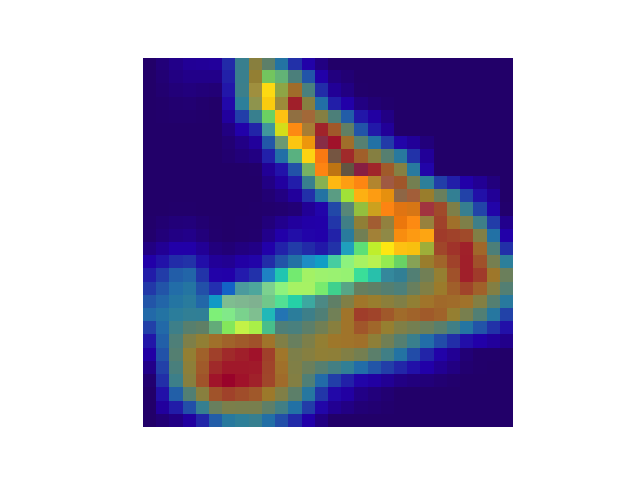}
    \includegraphics[width=0.22\linewidth]{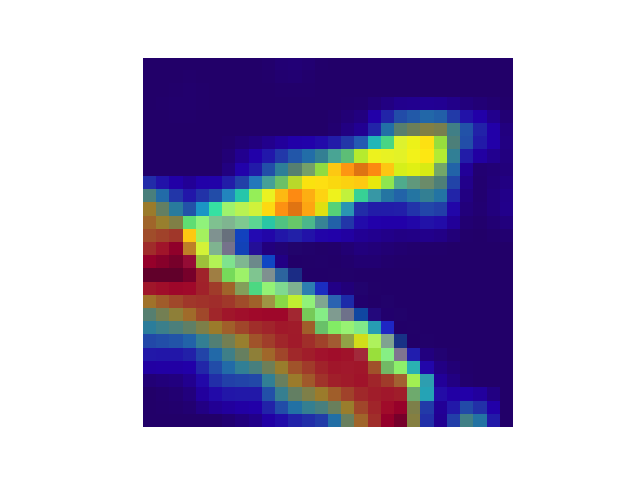}
    \includegraphics[width=0.22\linewidth]{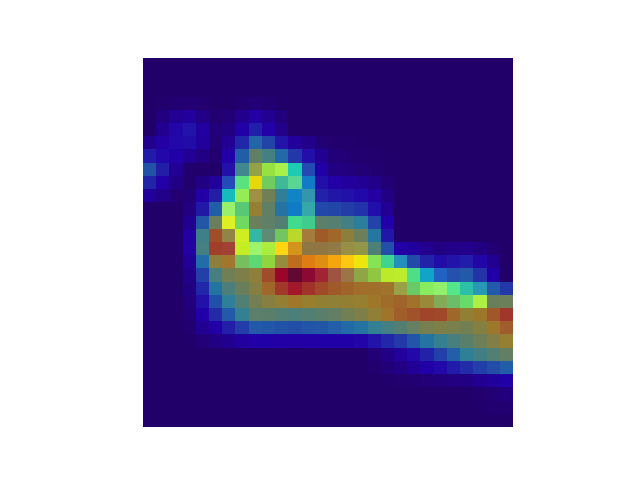}
    \caption{C-VAE Saliency Maps}
    \label{fig:cvae-saliency-maps}
\end{figure}

The results displayed in Figure \ref{fig:cvae-saliency-maps} and \ref{fig:cgan-saliency-maps} highlight the visual interpretation of the learning process our model goes through while performing digit classification, using a technique known as Saliency Maps \cite{simonyan2013deep} . These maps illuminate the regions in an image that are most salient or relevant to making a particular classification decision. In our case, the model has to recognize and classify different handwritten Arabic digits \cite{adebayo2018sanity}.

The Saliency Maps thus represent how our model "sees" and interprets these digits, effectively acting as a heatmap of model attentiveness. Bright regions in the Saliency Maps signify areas where small changes in the pixel values significantly affect the classification outcome, denoting high sensitivity. In contrast, darker regions imply areas of lower sensitivity \cite{moosmann2006learning}.

In the context of digit classification, these maps confirm that the unique, distinguishing attributes of each digit, such as the particular curves, edges, or other specific strokes that differentiate one digit from another, are the key factors influencing the correct identification and classification. These features appear to be the most salient or critical in the eyes of our model \cite{arun2021assessing}.

Notably, the successful recognition and highlighting of these unique digit attributes by the Saliency Maps further validate the high accuracy achieved by our Optical Character Recognition (OCR) model. It was trained on a dataset augmented by the Conditional Variational Autoencoder (C-VAE), as these maps effectively demonstrate that the model correctly focuses on the most important and distinguishing features in its learning process.

This outcome bolsters our confidence in the model's learning process, its understanding of the data, and its subsequent performance. It also provides us with an intuitive way to visually verify and interpret the decision-making process of the model, thereby reaffirming the robustness and reliability of the OCR model's high accuracy.

\subsection{Summary}
The proposed methodology focuses on optimizing OCR performance for Arabic handwritten digit recognition and examining the impact of synthetic images generated by GAN and VAE models on model accuracy and generalization capabilities. This comprehensive approach includes several steps, such as model compilation, data preprocessing, data augmentation, model training, validation, hyperparameter tuning, and evaluation on the test dataset.

By utilizing the LFID metric to assess the quality of synthetic images and evaluating the OCR model's performance on augmented datasets, the study aims to provide valuable insights into the effectiveness of various generation loss functions and data augmentation strategies. Furthermore, it investigates the correlation between LFID scores and OCR model accuracy, as well as comparing the most critical OCR features with unique features of Arabic handwritten digits.

This in-depth analysis helps identify the most promising techniques for improving OCR performance and reveals the benefits and limitations of different synthetic data generation approaches in the context of OCR tasks. Ultimately, the study contributes to a better understanding of the factors that influence OCR performance and offers guidance for practitioners seeking to develop more accurate and robust OCR models using synthetic data augmentation.

\section{Conclusion}
In conclusion, our study highlighted the advantages of Conditional Variational Autoencoders (C-VAEs) over Conditional Generative Adversarial Networks (C-GANs) for improving real-time OCR performance, especially in the context of Arabic handwritten digit recognition. The VAE model's superior performance can be attributed to its ability to generate synthetic images ten times faster than GANs, resulting in a more efficient training process.

A significant contribution of our research is the development of the \textbf{Synthetic Image Evaluation Procedure}(Algorithm \ref{alg:3}), an accurate and efficient alternative to traditional FID scores for monitoring the quality of synthetic images generated for OCR applications. The evaluation procedure enables real-time evaluation of model performance and supports early stopping during training, further optimizing the OCR system.

Our analysis, corroborated by Saliency Maps, validated the improvement in OCR performance. We demonstrated that the enhanced OCR system effectively leverages unique features of Arabic digits for classification, confirming the system identifies and classifies the digits based on their true unique features, rather than relying on irrelevant or trivial patterns.

By integrating generative data augmentation techniques like C-VAEs with innovative evaluation metrics such as LFID, our approach sets the stage for substantial advancements in OCR performance. These improvements are particularly valuable for real-time applications, where challenges such as noise, distortions, and limited training data availability often impede system accuracy. The insights derived from our research have the potential to guide the development of more advanced OCR systems capable of addressing a broader range of applications and adapting to various contexts.

Future work could explore other generative models and data augmentation techniques, as well as the application of our approach to other languages or domains. Additionally, further research could focus on improving the LFID metric, making it more robust and adaptable for different tasks and contexts.
\bibliographystyle{unsrt}  

\begin{thebibliography}{999}
	
	\bibitem{castleman1996digital}
	Kenneth~R Castleman.
	\newblock {\em {Digital image processing}}.
	\newblock Prentice Hall Press, 1996.
	
	\bibitem{niblack1985introduction}
	Wayne Niblack.
	\newblock {\em {An introduction to digital image processing}}.
	\newblock Strandberg Publishing Company, 1985.
	
	\bibitem{jain1989fundamentals}
	Anil~K Jain.
	\newblock {\em {Fundamentals of digital image processing}}.
	\newblock Prentice-Hall, Inc., 1989.
	
	\bibitem{jahne2005digital}
	Bernd J{\"{a}}hne.
	\newblock {\em {Digital image processing}}.
	\newblock Springer Science {\&} Business Media, 2005.
	
	\bibitem{boyat2015review}
	Ajay~Kumar Boyat and Brijendra~Kumar Joshi.
	\newblock {A review paper: noise models in digital image processing}.
	\newblock {\em arXiv preprint arXiv:1505.03489}, 2015.
	
	\bibitem{Al-Wzwazy2016HandwrittenNetworks}
	Haider Al-Wzwazy, Hayder Mohammed~Albehadili, Younes Saood~Alwan, Naz Islam,
	Haider~A Alwzwazy, Hayder~M Albehadili, Younes~S Alwan, Naz~E Islam, and
	Me~Student.
	\newblock {Handwritten Digit Recognition Using Convolutional Neural Networks}.
	\newblock {\em Article in International Journal of Innovative Research in
		Computer and Communication Engineering}, 3297(2), 2016.
	
	\bibitem{el2007two}
	Ezzat~Ali El-Sherif and Sherif Abdelazeem.
	\newblock {A Two-Stage System for Arabic Handwritten Digit Recognition Tested
		on a New Large Database.}
	\newblock In {\em Artificial intelligence and pattern recognition}, pages
	237--242, 2007.
	
	\bibitem{mori1999optical}
	Shunji Mori, Hirobumi Nishida, and Hiromitsu Yamada.
	\newblock {\em {Optical character recognition}}.
	\newblock John Wiley {\&} Sons, Inc., 1999.
	
	\bibitem{mithe2013optical}
	Ravina Mithe, Supriya Indalkar, and Nilam Divekar.
	\newblock {Optical character recognition}.
	\newblock {\em International journal of recent technology and engineering
		(IJRTE)}, 2(1):72--75, 2013.
	
	\bibitem{islam2017survey}
	Noman Islam, Zeeshan Islam, and Nazia Noor.
	\newblock {A survey on optical character recognition system}.
	\newblock {\em arXiv preprint arXiv:1710.05703}, 2017.
	
	\bibitem{chaudhuri2017optical}
	Arindam Chaudhuri, Krupa Mandaviya, Pratixa Badelia, Soumya K~Ghosh, Arindam
	Chaudhuri, Krupa Mandaviya, Pratixa Badelia, and Soumya~K Ghosh.
	\newblock {\em {Optical character recognition systems}}.
	\newblock Springer, 2017.
	
	\bibitem{singh2013optical}
	Sukhpreet Singh.
	\newblock {Optical character recognition techniques: a survey}.
	\newblock {\em Journal of emerging Trends in Computing and information
		Sciences}, 4(6), 2013.
	
	\bibitem{rao2016optical}
	N~Venkata Rao, ASCS Sastry, A~S~N Chakravarthy, and P~Kalyanchakravarthi.
	\newblock {OPTICAL CHARACTER RECOGNITION TECHNIQUE ALGORITHMS.}
	\newblock {\em Journal of Theoretical {\&} Applied Information Technology},
	83(2), 2016.
	
	\bibitem{berchmans2014optical}
	Deepa Berchmans and S~S Kumar.
	\newblock {Optical character recognition: an overview and an insight}.
	\newblock In {\em 2014 International Conference on Control, Instrumentation,
		Communication and Computational Technologies (ICCICCT)}, pages 1361--1365.
	IEEE, 2014.
	
	\bibitem{namysl2019efficient}
	Marcin Namysl and Iuliu Konya.
	\newblock {Efficient, lexicon-free OCR using deep learning}.
	\newblock In {\em 2019 international conference on document analysis and
		recognition (ICDAR)}, pages 295--301. IEEE, 2019.
	
	\bibitem{chernyshova2018generation}
	Yulia~S Chernyshova, Alexander~V Gayer, and Alexander~V Sheshkus.
	\newblock {Generation method of synthetic training data for mobile OCR system}.
	\newblock In {\em Tenth international conference on machine vision (ICMV
		2017)}, volume 10696, pages 640--646. SPIE, 2018.
	
	\bibitem{storchan2019data}
	Victor Storchan and Jocelyn Beauschene.
	\newblock {Data augmentation via adversarial networks for optical character
		recognition/conference submissions}.
	\newblock In {\em 2019 International Conference on Document Analysis and
		Recognition (ICDAR)}, pages 184--189. IEEE, 2019.
	
	\bibitem{mori1992historical}
	Shunji Mori, Ching~Y Suen, and Kazuhiko Yamamoto.
	\newblock {Historical review of OCR research and development}.
	\newblock {\em Proceedings of the IEEE}, 80(7):1029--1058, 1992.
	
	\bibitem{kolak2003generative}
	Okan Kolak, Bill Byrne, and Philip Resnik.
	\newblock {A generative probabilistic OCR model for NLP applications}.
	\newblock In {\em Proceedings of the 2003 Human Language Technology Conference
		of the North American Chapter of the Association for Computational
		Linguistics}, pages 134--141, 2003.
	
	\bibitem{kissos2016ocr}
	Ido Kissos and Nachum Dershowitz.
	\newblock {OCR error correction using character correction and feature-based
		word classification}.
	\newblock In {\em 2016 12th IAPR Workshop on Document Analysis Systems (DAS)},
	pages 198--203. IEEE, 2016.
	
	\bibitem{sporici2020improving}
	Dan Sporici, Elena Cușnir, and Costin-Anton Boiangiu.
	\newblock {Improving the accuracy of Tesseract 4.0 OCR engine using
		convolution-based preprocessing}.
	\newblock {\em Symmetry}, 12(5):715, 2020.
	
	\bibitem{liu2023real}
	Weijia Liu, Jiuxin Cao, Yilin Zhu, Bo~Liu, and Xuelin Zhu.
	\newblock {Real-time anomaly detection on surveillance video with two-stream
		spatio-temporal generative model}.
	\newblock {\em Multimedia systems}, 29(1):59--71, 2023.
	
	\bibitem{tkach2016sphere}
	Anastasia Tkach, Mark Pauly, and Andrea Tagliasacchi.
	\newblock {Sphere-meshes for real-time hand modeling and tracking}.
	\newblock {\em ACM Transactions on Graphics (ToG)}, 35(6):1--11, 2016.
	
	\bibitem{taghva1996evaluation}
	Kazem Taghva, Julie Borsack, and Allen Condit.
	\newblock {Evaluation of model-based retrieval effectiveness with OCR text}.
	\newblock {\em ACM Transactions on Information Systems (TOIS)}, 14(1):64--93,
	1996.
	
	\bibitem{blue1994evaluation}
	James~L Blue, Gerald~T Candela, Patrick~J Grother, Rama Chellappa, and
	Charles~L Wilson.
	\newblock {Evaluation of pattern classifiers for fingerprint and OCR
		applications}.
	\newblock {\em Pattern Recognition}, 27(4):485--501, 1994.
	
	\bibitem{cai2020real}
	Qing Cai, Mohamed Abdel-Aty, Jinghui Yuan, Jaeyoung Lee, and Yina Wu.
	\newblock {Real-time crash prediction on expressways using deep generative
		models}.
	\newblock {\em Transportation research part C: emerging technologies},
	117:102697, 2020.
	
	\bibitem{fasel2005generative}
	Ian Fasel, Bret Fortenberry, and Javier Movellan.
	\newblock {A generative framework for real time object detection and
		classification}.
	\newblock {\em Computer Vision and Image Understanding}, 98(1):182--210, 2005.
	
	\bibitem{Goodfellow2014}
	Ian Goodfellow, Jean Pouget-Abadie, Mehdi Mirza, Bing Xu, David Warde-Farley,
	Sherjil Ozair, and Yoshua Bengio.
	\newblock {Generative adversarial nets}.
	\newblock In {\em Advances in Neural Information Processing Systems},
	volume~27, pages 2672--2680, 2014.
	
	\bibitem{bond2021deep}
	Sam Bond-Taylor, Adam Leach, Yang Long, and Chris~G Willcocks.
	\newblock {Deep Generative Modelling: A Comparative Review of VAEs, GANs,
		Normalizing Flows, Energy-Based and Autoregressive Models}.
	\newblock {\em IEEE Transactions on Pattern Analysis and Machine Intelligence},
	44(11):7327--7347, 2021.
	
	\bibitem{kingma2019introduction}
	Diederik~P Kingma and Max Welling.
	\newblock {\em {An Introduction to Variational Autoencoders}}, volume~12.
	\newblock Now Publishers, Inc., 2019.
	
	\bibitem{girin2020dynamical}
	Laurent Girin, Simon Leglaive, Xiaoyu Bie, Julien Diard, Thomas Hueber, and
	Xavier Alameda-Pineda.
	\newblock {Dynamical variational autoencoders: A comprehensive review}.
	\newblock {\em arXiv preprint arXiv:2008.12595}, 2020.
	
	\bibitem{Sohn2015CGAN}
	Kihyuk Sohn, Honglak Lee, and Xinchen Yan.
	\newblock {Learning Structured Output Representation using Deep Conditional
		Generative Models}.
	\newblock {\em Advances in Neural Information Processing Systems}, 28, 2015.
	
	\bibitem{Mirza2014}
	Mehdi Mirza and Simon Osindero.
	\newblock {Conditional Generative Adversarial Nets}.
	\newblock {\em Technical report}, 11 2014.
	
	\bibitem{lim2018molecular}
	Jaechang Lim, Seongok Ryu, Jin~Woo Kim, and Woo~Youn Kim.
	\newblock {Molecular generative model based on conditional variational
		autoencoder for de novo molecular design}.
	\newblock {\em Journal of cheminformatics}, 10(1):1--9, 2018.
	
	\bibitem{asperti2020balancing}
	Andrea Asperti and Matteo Trentin.
	\newblock {Balancing reconstruction error and kullback-leibler divergence in
		variational autoencoders}.
	\newblock {\em IEEE Access}, 8:199440--199448, 2020.
	
	\bibitem{aggarwal2021generative}
	Alankrita Aggarwal, Mamta Mittal, and Gopi Battineni.
	\newblock {Generative adversarial network: An overview of theory and
		applications}.
	\newblock {\em International Journal of Information Management Data Insights},
	1(1):100004, 2021.
	
	\bibitem{cetin2023attri}
	Irem Cetin, Maialen Stephens, Oscar Camara, and Miguel A~González Ballester.
	\newblock {Attri-VAE: Attribute-based interpretable representations of medical
		images with variational autoencoders}.
	\newblock {\em Computerized Medical Imaging and Graphics}, 104:102158, 2023.
	
	\bibitem{doermann2014handbook}
	David Doermann and Karl Tombre.
	\newblock {\em {Handbook of document image processing and recognition}}.
	\newblock Springer Publishing Company, Incorporated, 2014.
	
	\bibitem{cai2019multi}
	Lei Cai, Hongyang Gao, and Shuiwang Ji.
	\newblock {Multi-stage variational auto-encoders for coarse-to-fine image
		generation}.
	\newblock In {\em Proceedings of the 2019 SIAM International Conference on Data
		Mining}, pages 630--638. SIAM, 2019.
	
	\bibitem{kebaili2023deep}
	Aghiles Kebaili, Jérôme Lapuyade-Lahorgue, and Su~Ruan.
	\newblock {Deep Learning Approaches for Data Augmentation in Medical Imaging: A
		Review}.
	\newblock {\em Journal of Imaging}, 9(4):81, 2023.
	
	\bibitem{Kingma2013}
	Diederik~P Kingma and Max Welling.
	\newblock {Auto-encoding variational Bayes}.
	\newblock In {\em International Conference on Learning Representations (ICLR)},
	2013.
	
	\bibitem{Zhu2017Cycle}
	Jun-Yan~Yan Zhu, Taesung Park, Phillip Isola, and Alexei~A. Efros.
	\newblock {Unpaired Image-to-Image Translation using Cycle-Consistent
		Adversarial Networks}.
	\newblock {\em Proceedings of the IEEE International Conference on Computer
		Vision}, 2017-Octob, 3 2017.
	
	\bibitem{karras2019style}
	Tero Karras, Samuli Laine, and Timo Aila.
	\newblock {A style-based generator architecture for generative adversarial
		networks}.
	\newblock In {\em Proceedings of the IEEE/CVF conference on computer vision and
		pattern recognition}, pages 4401--4410, 2019.
	
	\bibitem{brock2018large}
	Andrew Brock, Jeff Donahue, and Karen Simonyan.
	\newblock {Large Scale GAN Training for High Fidelity Natural Image Synthesis}.
	\newblock {\em arXiv preprint arXiv:1809.11096}, 9 2018.
	
	\bibitem{barratt2018note}
	Shane Barratt and Rishi Sharma.
	\newblock {A Note on the Inception Score}.
	\newblock {\em arXiv preprint arXiv:1801.01973}, 2018.
	
	\bibitem{contreras2012kullback}
	Javier~E Contreras-Reyes and Reinaldo~B Arellano-Valle.
	\newblock {Kullback--Leibler divergence measure for multivariate skew-normal
		distributions}.
	\newblock {\em Entropy}, 14(9):1606--1626, 2012.
	
	\bibitem{menendez1997jensen}
	M~L Men{\'{e}}ndez, J~A Pardo, L~Pardo, and M~C Pardo.
	\newblock {The jensen-shannon divergence}.
	\newblock {\em Journal of the Franklin Institute}, 334(2):307--318, 1997.
	
	\bibitem{vallender1974calculation}
	S~S Vallender.
	\newblock {Calculation of the Wasserstein distance between probability
		distributions on the line}.
	\newblock {\em Theory of Probability {\&} Its Applications}, 18(4):784--786,
	1974.
	
	\bibitem{borgwardt2006integrating}
	Karsten~M Borgwardt, Arthur Gretton, Malte~J Rasch, Hans-Peter Kriegel,
	Bernhard Sch{\"{o}}lkopf, and Alex~J Smola.
	\newblock {Integrating structured biological data by kernel maximum mean
		discrepancy}.
	\newblock {\em Bioinformatics}, 22(14):e49--e57, 2006.
	
	\bibitem{Heusel2017}
	Martin Heusel, Hubert Ramsauer, Thomas Unterthiner, Bernhard Nessler, and Sepp
	Hochreiter.
	\newblock {GANs trained by a two time-scale update rule converge to a local
		Nash equilibrium}.
	\newblock {\em Advances in Neural Information Processing Systems},
	2017-Decem(Nips):6627--6638, 2017.
	
	\bibitem{ahmed2016improving}
	Md~Sajib Ahmed, Teresa Gon{\c{c}}alves, and Hasan Sarwar.
	\newblock {Improving Bangla OCR output through correction algorithms}.
	\newblock In {\em 2016 10th International Conference on Software, Knowledge,
		Information Management {\&} Applications (SKIMA)}, pages 338--343. IEEE,
	2016.
	
	\bibitem{apuke2017quantitative}
	Oberiri~Destiny Apuke.
	\newblock {Quantitative research methods: A synopsis approach}.
	\newblock {\em Kuwait Chapter of Arabian Journal of Business and Management
		Review}, 33(5471):1--8, 2017.
	
	\bibitem{Sawy2017}
	Ahmed~El Sawy, Hazem El-bakry, Mohamed Loey, and Hazem El-bakry.
	\newblock {CNN for handwritten arabic digits recognition based on LeNet-5}.
	\newblock {\em Advances in Intelligent Systems and Computing}, 533:565--575,
	2017.
	
	\bibitem{cubuk2018autoaugment}
	Ekin~D Cubuk, Barret Zoph, Dandelion Mane, Vijay Vasudevan, and Quoc~V Le.
	\newblock {Autoaugment: Learning augmentation policies from data}.
	\newblock {\em arXiv preprint arXiv:1805.09501}, 2018.
	
	\bibitem{frid2018synthetic}
	Maayan Frid-Adar, Eyal Klang, Michal Amitai, Jacob Goldberger, and Hayit
	Greenspan.
	\newblock {Synthetic data augmentation using GAN for improved liver lesion
		classification}.
	\newblock In {\em 2018 IEEE 15th international symposium on biomedical imaging
		(ISBI 2018)}, pages 289--293. IEEE, 2018.
	
	\bibitem{amari1993backpropagation}
	Shun-ichi Amari.
	\newblock {Backpropagation and stochastic gradient descent method}.
	\newblock {\em Neurocomputing}, 5(4-5):185--196, 1993.
	
	\bibitem{ho2019real}
	Yaoshiang Ho and Samuel Wookey.
	\newblock {The real-world-weight cross-entropy loss function: Modeling the
		costs of mislabeling}.
	\newblock {\em IEEE access}, 8:4806--4813, 2019.
	
	\bibitem{kim2021conditional}
	Jaehyeon Kim, Jungil Kong, and Juhee Son.
	\newblock {Conditional variational autoencoder with adversarial learning for
		end-to-end text-to-speech}.
	\newblock In {\em International Conference on Machine Learning}, pages
	5530--5540. PMLR, 2021.
	
	\bibitem{pang2020deep}
	Bo~Pang, Erik Nijkamp, and Ying~Nian Wu.
	\newblock {Deep learning with tensorflow: A review}.
	\newblock {\em Journal of Educational and Behavioral Statistics},
	45(2):227--248, 2020.
	
	\bibitem{zhang2018improved}
	Zijun Zhang.
	\newblock {Improved adam optimizer for deep neural networks}.
	\newblock In {\em 2018 IEEE/ACM 26th international symposium on quality of
		service (IWQoS)}, pages 1--2. Ieee, 2018.
	
	\bibitem{bock2018improvement}
	Sebastian Bock, Josef Goppold, and Martin Wei{\ss}.
	\newblock {An improvement of the convergence proof of the ADAM-Optimizer}.
	\newblock {\em arXiv preprint arXiv:1804.10587}, 2018.
	
	\bibitem{johnson2016structured}
	Matthew~J Johnson, David Duvenaud, Alexander~B Wiltschko, Sandeep~R Datta, and
	Ryan~P Adams.
	\newblock {Structured VAEs: Composing probabilistic graphical models and
		variational autoencoders}.
	\newblock {\em arXiv preprint arXiv:1603.06277}, 2:2016, 2016.
	
	\bibitem{sonderby2016ladder}
	Casper~Kaae S{\o}nderby, Tapani Raiko, Lars Maal{\o}e, Søren~Kaae S{\o}nderby,
	and Ole Winther.
	\newblock {Ladder variational autoencoders}.
	\newblock {\em Advances in neural information processing systems}, 29, 2016.
	
	\bibitem{yang2017improved}
	Zichao Yang, Zhiting Hu, Ruslan Salakhutdinov, and Taylor Berg-Kirkpatrick.
	\newblock {Improved variational autoencoders for text modeling using dilated
		convolutions}.
	\newblock In {\em International conference on machine learning}, pages
	3881--3890. PMLR, 2017.
	
	\bibitem{kingma2014adam}
	Diederik~P Kingma and Jimmy~Lei Ba.
	\newblock {Adam: A Method for Stochastic Optimization}.
	\newblock {\em 3rd International Conference on Learning Representations, ICLR
		2015 - Conference Track Proceedings}, 2014.
	
	\bibitem{sarika2021cnn}
	Naragudem Sarika, Nageswararao Sirisala, and Muni~Sekhar Velpuru.
	\newblock {CNN based optical character recognition and applications}.
	\newblock In {\em 2021 6th International conference on inventive computation
		technologies (ICICT)}, pages 666--672. IEEE, 2021.
	
	\bibitem{alt1995computing}
	Helmut Alt and Michael Godau.
	\newblock {COMPUTING THE FR{\'{E}}CHET DISTANCE BETWEEN TWO POLYGONAL CURVES}.
	\newblock {\em International Journal of Computational Geometry
		{\{}{\textbackslash}textbackslash{\}}{\&} Applications}, 1995.
	
	\bibitem{har2014frechet}
	Sariel Har-Peled and Benjamin Raichel.
	\newblock {The Fr{\'{e}}chet distance revisited and extended}.
	\newblock {\em ACM Transactions on Algorithms (TALG)}, 10(1):1--22, 2014.
	
	\bibitem{szegedy2016rethinking}
	Christian Szegedy, Vincent Vanhoucke, Sergey Ioffe, Jon Shlens, and Zbigniew
	Wojna.
	\newblock {Rethinking the inception architecture for computer vision}.
	\newblock In {\em Proceedings of the IEEE conference on computer vision and
		pattern recognition}, pages 2818--2826, 2016.
	
	\bibitem{Szegedy2015deeper}
	Christian Szegedy, Wei Liu, Yangqing Jia, Pierre Sermanet, Scott Reed, Dragomir
	Anguelov, Dumitru Erhan, Vincent Vanhoucke, and Andrew Rabinovich.
	\newblock {Going deeper with convolutions}.
	\newblock {\em IEEE conference on computer vision and pattern recognition.}, 9
	2015.
	
	\bibitem{olah2017feature}
	Chris Olah, Alexander Mordvintsev, and Ludwig Schubert.
	\newblock {Feature visualization}.
	\newblock {\em Distill}, 2(11):e7, 2017.
	
	\bibitem{kohler2012feedback}
	Barbara K{\"{o}}hler, Juan Haladjian, Blagina Simeonova, and Damir
	Ismailovi{\'{c}}.
	\newblock {Feedback in low vs. high fidelity visuals for game prototypes}.
	\newblock In {\em 2012 Second International Workshop on Games and Software
		Engineering: Realizing User Engagement with Game Engineering Techniques
		(GAS)}, pages 42--47. IEEE, 2012.
	
	\bibitem{jung2021internalized}
	Steffen Jung and Margret Keuper.
	\newblock {Internalized biases in fr{\'{e}}chet inception distance}.
	\newblock In {\em NeurIPS 2021 Workshop on Distribution Shifts: Connecting
		Methods and Applications}, 2021.
	
	\bibitem{yu2021frechet}
	Yu~Yu, Weibin Zhang, and Yun Deng.
	\newblock {Frechet inception distance (fid) for evaluating gans}.
	\newblock {\em China University of Mining Technology Beijing Graduate School:
		Beijing, China}, 2021.
	
	\bibitem{Krizhevsky2012}
	Alex Krizhevsky, Ilya Sutskever, and Geoffrey~E Hinton.
	\newblock {ImageNet classification with deep convolutional neural networks}.
	\newblock {\em Advances in Neural Information Processing Systems},
	25:1097--1105, 2012.
	
	\bibitem{yao2007early}
	Yuan Yao, Lorenzo Rosasco, and Andrea Caponnetto.
	\newblock {On early stopping in gradient descent learning}.
	\newblock {\em Constructive Approximation}, 26:289--315, 2007.
	
	\bibitem{bilbrey2020look}
	Jenna~A Bilbrey, Joseph~P Heindel, Malachi Schram, Pradipta Bandyopadhyay,
	Sotiris~S Xantheas, and Sutanay Choudhury.
	\newblock {A look inside the black box: Using graph-theoretical descriptors to
		interpret a Continuous-Filter Convolutional Neural Network (CF-CNN) trained
		on the global and local minimum energy structures of neutral water clusters}.
	\newblock {\em The Journal of Chemical Physics}, 153(2), 2020.
	
	\bibitem{adebayo2018sanity}
	Julius Adebayo, Justin Gilmer, Michael Muelly, Ian Goodfellow, Moritz Hardt,
	and Been Kim.
	\newblock {Sanity checks for saliency maps}.
	\newblock {\em Advances in neural information processing systems}, 31, 2018.
	
	\bibitem{sharma2012discriminative}
	Gaurav Sharma, Frédéric Jurie, and Cordelia Schmid.
	\newblock {Discriminative spatial saliency for image classification}.
	\newblock In {\em 2012 IEEE Conference on Computer Vision and Pattern
		Recognition}, pages 3506--3513. IEEE, 2012.
	
	\bibitem{moayeri2022comprehensive}
	Mazda Moayeri, Phillip Pope, Yogesh Balaji, and Soheil Feizi.
	\newblock {A comprehensive study of image classification model sensitivity to
		foregrounds, backgrounds, and visual attributes}.
	\newblock In {\em Proceedings of the IEEE/CVF Conference on Computer Vision and
		Pattern Recognition}, pages 19087--19097, 2022.
	
	\bibitem{simonyan2013deep}
	Karen Simonyan, Andrea Vedaldi, and Andrew Zisserman.
	\newblock {Deep inside convolutional networks: Visualising image classification
		models and saliency maps}.
	\newblock {\em arXiv preprint arXiv:1312.6034}, 2013.
	
	\bibitem{moosmann2006learning}
	Franck Moosmann, Diane Larlus, and Frederic Jurie.
	\newblock {Learning saliency maps for object categorization}.
	\newblock In {\em International Workshop on The Representation and Use of Prior
		Knowledge in Vision (in ECCV'06)}, 2006.
	
	\bibitem{arun2021assessing}
	Nishanth Arun, Nathan Gaw, Praveer Singh, Ken Chang, Mehak Aggarwal, Bryan
	Chen, Katharina Hoebel, Sharut Gupta, Jay Patel, Mishka Gidwani, and
	{others}.
	\newblock {Assessing the trustworthiness of saliency maps for localizing
		abnormalities in medical imaging}.
	\newblock {\em Radiology: Artificial Intelligence}, 3(6):e200267, 2021.
	
\end{thebibliography}

\end{document}